\documentclass{article}

\usepackage{arxiv}

\newacronym{em}{EM}{electromagnetic}
\newacronym[plural=LLMs,longplural=large language models]{llm}{LLM}{large language model}
\newacronym[plural=CPUs,longplural=central processing units]{cpu}{CPU}{central processing unit}
\newacronym[plural=GPUs,longplural=graphics processing units]{gpu}{GPU}{graphics processing unit}
\newacronym{ad}{AD}{automatic differentiation}
\newacronym{ml}{ML}{machine learning}
\newacronym{mcp}{MCP}{Model Context Protocol}
\newacronym{rag}{RAG}{retrieval-augmented generation}
\newacronym[plural=NNs,longplural=neural networks]{nn}{NN}{neural network}
\newacronym[plural=APIs,longplural=application programming interfaces]{api}{API}{application programming interface}
\newacronym{des}{DES}{design efficiency score}
\newacronym{ai}{AI}{artificial intelligence}
\newacronym{rcwa}{RCWA}{rigorous coupled-wave analysis}
\newacronym{rdit}{R-DIT}{rigorous diffraction interface theory}
\newacronym{fmm}{FMM}{Fourier modal method}
\newacronym{midir}{mid-IR}{mid-infrared}

\newacronym{mce}{MCE}{meta context engineering}
\newcommand{\SameType}{Same-type\xspace}
\newcommand{\NewTypeA}{New-type-A\xspace}
\newcommand{\NewTypeB}{New-type-B\xspace}
\newacronym{sg}{SG}{success goal}
\newacronym{se_rate}{SE}{success execution}
\newacronym{cpf}{CPF}{criteria pass fraction}
\newacronym{bm}{BM}{bottleneck margin}

\newcommand{\solver}{\texttt{TorchRDIT}}

\title{A Self-Evolving Agentic Framework for Metasurface Inverse Design}

\author[1]{Yi Huang\,\orcidlink{0000-0002-8399-3225}\,\textsuperscript{*,}}
\author[1]{Bowen Zheng}
\author[1]{Yunxi Dong}
\author[1]{Hong Tang}
\author[1]{Huan Zhao}
\author[1]{S. M. Rakibul Hasan Shawon}
\author[1]{Hualiang Zhang,\textsuperscript{\#,}}
\affil[1]{Department of Electrical and Computer Engineering, University of Massachusetts Lowell}
\affil[ ]{\textsuperscript{*}\texttt{Yi\_Huang@student.uml.edu}, \textsuperscript{\#}\texttt{Hualiang\_Zhang@uml.edu}}

\date{}

\renewcommand{\headeright}{}
\renewcommand{\undertitle}{}
\renewcommand{\shorttitle}{}

\hypersetup{
pdftitle={A Self-Evolving Agentic Framework for Metasurface Inverse Design},
pdfsubject={metasurface inverse design, self-evolving agentic framework},
pdfauthor={Yi Huang, Hualiang Zhang},
pdfkeywords={metasurface inverse design, self-evolving agents, agentic workflow, skill evolution, large language model (LLM), differentiable electromagnetic solver},
}

\begin{document}
\maketitle

\begin{abstract}
Metasurface inverse design can realize complex optical functionality, but turning a target optical response into executable optimization code still requires substantial expertise in computational electromagnetics and solver-specific software engineering. We present a self-evolving agentic framework that lowers this barrier by coupling a coding agent, explicit human-readable skill files, and a deterministic physics-based evaluator. Rather than updating model weights, it revises the skill files from solver-grounded feedback, while the base model and differentiable solver, which provides the physics simulation and gradients, stay fixed. On a multi-type benchmark, skill evolution raises same-type task success from 38\% to 74\%, the fraction of physical criteria met from 0.51 to 0.87, and reduces average attempts from 4.10 to 2.30. On two new-type families, success holds near ceiling on one (0.92 to 0.90) and rises from 0.20 to 0.90 on the other. Skill evolution offers a practical path toward autonomous and accessible inverse-design workflows.
\end{abstract}


\section{Introduction}
\label{sec:agentic-introduction}

Metasurface design is increasingly cast as an inverse problem, in which a target optical response is specified and the corresponding structure is sought \cite{jensenTopologyOptimizationNanophotonics2011, moleskyInverseDesignNanophotonics2018}. This paradigm reflects the growing demand for broadband, multifunctional, and highly integrated meta-optical devices, whose design spaces are high-dimensional, strongly coupled, and resistant to intuition-guided exploration, particularly under freeform parameterizations \cite{chenReviewMetasurfacesPhysics2016, ouAdvancesMetaOpticsMetasurfaces2023, parkFreeformOptimizationNanophotonic2022, liEmpoweringMetasurfacesInverse2022, huang2024differentiable, huangEigendecompositionfreeInverseDesign2024}. In parallel, deep-learning-assisted methods have broadened the computational design landscape for metasurfaces and flat optics \cite{dongAdvancedDeepLearning2025, gaoInverseDesignFlat2022, dongAchromaticSingleMetalens2024}. As a result, manual geometry tuning and parameter sweeps no longer suffice to identify high-performance structures, and computational inverse-design has become central to metasurface research \cite{jensenTopologyOptimizationNanophotonics2011, moleskyInverseDesignNanophotonics2018, liEmpoweringMetasurfacesInverse2022, huangInverseDesignPhotonic2023, huang3DPrintedMillimeterWaveFreeForm2024, huangEigendecompositionfreeInverseDesign2024}. Yet this growing centrality has not translated into commensurate ease of automation: converting a desired optical response into a working design program still requires specialized knowledge of computational electromagnetics together with the solver-specific software engineering needed to implement and maintain it \cite{moleskyInverseDesignNanophotonics2018, liEmpoweringMetasurfacesInverse2022}. These barriers often confine inverse design to relatively narrow, task-specific design problems that are difficult to extend to more complex functionalities or to the broader multiphysics applications that interdisciplinary research requires.

Concretely, the researcher must select a parameterization, define the objective and constraints, decide how to initialize and optimize the variables, and verify the solver outputs \cite{moleskyInverseDesignNanophotonics2018, liEmpoweringMetasurfacesInverse2022}. Recent progress in \gls{llm}-based coding, tool use, and natural-language interaction makes part of this workflow easier. Early studies in nanophotonic inverse design mostly used \glspl{llm} as surrogate predictors or natural-language interfaces to existing tasks \cite{maOptoGPTFoundationModel2024, luLearningElectromagneticMetamaterial2025, kimNanophotonicDeviceDesign2025, zhangChatChipLarge2025}. More recent systems instead use the model to write code, call tools, and assemble optimization loops around established computational engines \cite{luAgenticFrameworkAutonomous2025, lupoiuMultiagenticFrameworkRealtime2025a}. Our recent \gls{mcp}-based framework showed that this workflow can be grounded in standardized solver access without compromising mathematical rigor \cite{huangMCPenabledLLMMetaoptics2025}. Even so, these systems can complete a task-specific workflow but do not retain solver-specific tactics in a form that later tasks can reuse \cite{luAgenticFrameworkAutonomous2025, lupoiuMultiagenticFrameworkRealtime2025a, huangMCPenabledLLMMetaoptics2025}.

Addressing that limitation requires a workflow that can keep and reuse what worked on earlier tasks. In this work, we evolve a set of explicit, human-readable skill files rather than model weights, revising them from per-task feedback. This follows the bi-level view of \gls{mce} \cite{yeMetaContextEngineering2026}, in which an outer loop revises the textual guidance used by an inner task-solving agent (an \gls{llm} that writes and runs code toward a goal). This keeps adaptation outside the base model, in line with recent agent frameworks that improve by updating their prompts and memory rather than through gradient updates \cite{fangComprehensiveSurveySelfEvolving2025, gaoSurveySelfEvolvingAgents2025, heEvoTestEvolutionaryTestTime2026, zhangAgenticContextEngineering2025}, and avoids the data demands and forgetting risks associated with fine-tuning \cite{luoEmpiricalStudyCatastrophic2025}. A separate deterministic evaluator scores each candidate with physical simulation rather than model self-assessment, so the base model, the solver, and the scoring rule remain fixed throughout the study \cite{openai_harness_2025, schmid_agent_harness_2026}. We therefore ask two concrete questions: whether skill evolution improves task success and search efficiency in physics-based metasurface inverse design, and whether the resulting guidance transfers to new-type task families, that is, task types not seen during skill development. To our knowledge, this is the first study to examine the evolution of skills in agentic metasurface inverse-design workflows.

Section~\ref{sec:agentic-methods} defines the framework and evaluation protocol. Section~\ref{sec:agentic-experiments} reports the experimental results on task success, efficiency, error patterns, agent behavior, cost, and transfer. Section~\ref{sec:agentic-discussion} discusses the implications and limits of the current framework.

\section{Methods}
\label{sec:agentic-methods}

\subsection{Framework Overview}
\label{subsec:framework-overview}

\begin{figure}[htb!]
	\centering
	\includegraphics[width=0.98\textwidth]{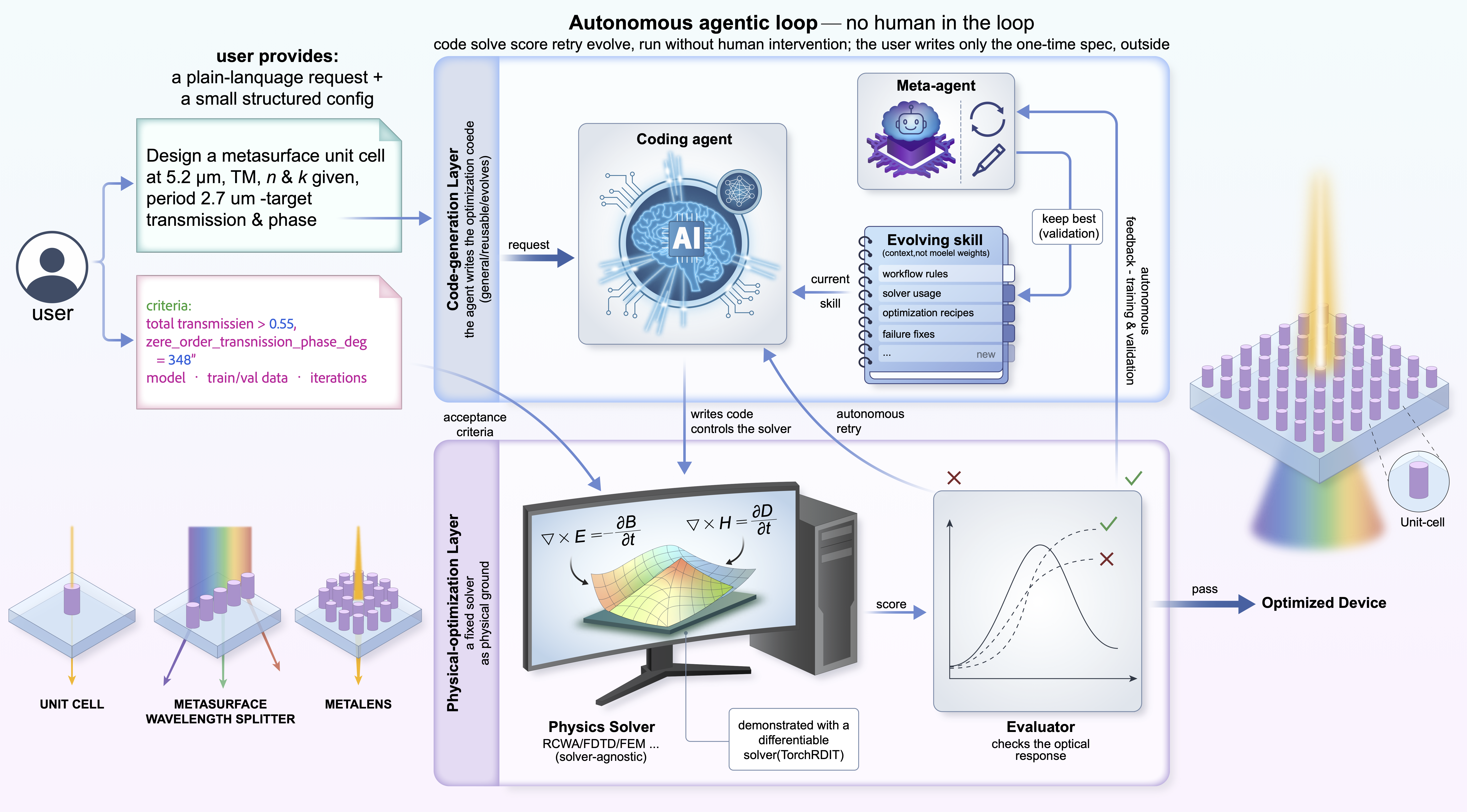}
	\caption{\textbf{Overview of the self-evolving agentic framework for metasurface inverse design.} The framework is organized as two decoupled layers: an upper code-generation layer, where a coding agent turns a natural-language request into a candidate optimization program guided by an evolving skill file (\texttt{SKILL.md}), and a lower physics layer, where a fixed differentiable solver performs the electromagnetic forward simulation and provides gradients, and an evaluator scores the result against the task criteria. The agent programs the fixed solver rather than replacing it, and the loop runs autonomously without a human in the loop. An inner loop retries the program when an evaluation fails, while an outer loop lets a meta-agent revise the skill file from per-task feedback, keeping the best version on validation; a passing design is released as the realized metasurface. The demonstrated solver is a differentiable RCWA implementation, {\solver}~\cite{huangEigendecompositionfreeInverseDesign2024}.}
	\label{fig:sch_agent-framework}
\end{figure}

The framework adds a code-generation layer on top of a fixed, differentiable solver, {\solver} \cite{huangEigendecompositionfreeInverseDesign2024}, which performs all electromagnetic forward simulations and, being differentiable, provides gradients of the optical response with respect to the design parameters. As shown in Figure~\ref{fig:sch_agent-framework}, this layer has three components: a meta-agent that revises the skill files, a coding agent that writes the optimization program for each task, and a deterministic evaluator that scores the result. Once a skill-evolution iteration finishes, the meta-agent inspects the per-task records and validation summaries to revise the current skill files. For a new task, the coding agent uses those files together with the task specification to write a candidate optimization program: it sets the design parameterization and the loss, and writes the optimization procedure itself, typically a conventional gradient-based loop that calls the solver at each step, though it may also use a derivative-free search. The evaluator executes that program with the {\solver}, checks the simulated design against the task criteria, and returns execution status, per-criterion outcomes, \gls{cpf}, \gls{bm}, and \gls{sg}. Because the solver is never modified, gains in task success reflect better optimization code written by the agent, not a change to the physics or the optimizer. The inputs required from the user are correspondingly minimal: for each task, the user provides only a natural-language design request specifying the target optical response and a set of acceptance criteria in a structured, machine-checkable format, and writes no optimization code, gradient or adjoint derivation, trained network, or labeled example designs. The per-family template files used to assemble the benchmark are a one-time cost of constructing the evaluation suite rather than a requirement for solving a new task. Those records are kept for later skill-evolution iterations, where they provide the evidence used to update the skills rather than relying on the short-lived chat history of a single task. Further implementation details are provided in Appendix~\ref{app:agentic-skill-artifact}.

Skill evolution follows the bi-level procedure \cite{yeMetaContextEngineering2026}, with skill revision driven directly by evaluator outputs on inverse-design tasks. At each iteration, the meta-agent reads per-task records from the training split, including execution errors, per-criterion feedback, \gls{cpf}, \gls{bm}, \gls{sg}, and attempt counts, and uses these records to revise the skill file.  The coding agent then applies the updated skill file to the next batch of training tasks. Candidate skill files are evaluated on a held-out validation split, and the best skill file is selected by the validation \gls{sg} rate:
\begin{equation}
    s^* = \arg\max_{s \in \mathcal{S}}\ \text{SG}_{\text{val}}(s),
    \label{eq:skill-selection}
\end{equation}
where $s$ denotes a candidate skill file and $\mathcal{S}$ is the collection of candidate skill files produced across iterations. Because candidates frequently tie on validation \gls{sg}, the maximization in Equation~\ref{eq:skill-selection} is resolved by a fixed secondary ordering: among files of equal validation \gls{sg}, the loop prefers higher \gls{cpf}, then smaller total constraint violation, then larger \gls{bm}, with \gls{se_rate}, solver cost, and finally the later iteration as remaining tiebreakers. The rule is elitist: the incumbent best is replaced only by a strictly higher-ranked candidate, so a revision that ranks no better on validation is never deployed. Further details of the generic skill-evolution procedure, including the skill-combination step that merges features from multiple candidate skill files, are given in the original work \cite{yeMetaContextEngineering2026}; Algorithm~\ref{alg:mce-inverse-design} summarizes the inverse-design instantiation used in this work.

\begin{algorithm}[t]
\caption{Per-task code generation and evaluation with two-level per-task retry (restarts and attempts).}
\label{alg:mce-inverse-design}
\begin{algorithmic}[1]
\Require task specification $x$, skill file $s$, EM solver $\mathcal{F}$, restarts $R$, attempts $A$
\Statex \textit{Called by the skill-evolution loop (the outer iteration; one iteration revises the skill file).}
\Statex \textit{Within a task, restarts reset the agent session and attempts revise the program in place.}
\Function{CodegenEval}{$x, s, \mathcal{F}, R, A$}
    \State $\text{best} \leftarrow \textsc{null}$
    \For{$r = 1, \ldots, R$} \Comment{restart: reset coding-agent session}
        \State $\text{restart\_best} \leftarrow \textsc{null}$
        \State initialize new agent session with skill $s$ and task $x$
        \If{$r > 1$} seed with best prior candidate and restart-level feedback
        \EndIf
        \For{$a = 1, \ldots, A$} \Comment{attempt: follow-up retry within the session}
            \State $\hat{y} \leftarrow f_\theta(x, s)$ \Comment{agent generates or revises the candidate program}
            \State $\mathbf{r} \leftarrow \Call{Execute}{\hat{y}, \mathcal{F}}$ \Comment{run script, invoke solver, collect solver results}
            \If{execution error}
                \State return feedback (error type, traceback) to agent
                \State \textbf{continue} \Comment{next attempt}
            \EndIf
            \State evaluate $\mathbf{r}$ with the fixed criteria scorer; compute per-criterion margins, SG, CPF, BM
            \State update $\text{restart\_best}$ if the current candidate outranks it by CPF, then BM
            \If{$\text{SG}(\mathbf{r}) = 1$} \Comment{stop: all criteria satisfied}
                \State \Return $(\text{SG}{=}1,\, \text{CPF},\, \text{BM},\, \text{attempts}{=}(r{-}1)A{+}a)$
            \EndIf
            \State return per-criterion margins to agent \Comment{next attempt}
        \EndFor
        \State update $\text{best}$ if $\text{restart\_best}$ outranks it by CPF, then BM
    \EndFor
    \If{$\text{best} = \textsc{null}$}
        \State \Return \textsc{ExecutionFailure}$(\text{attempts}{=}R \times A)$
    \EndIf
    \State \Return $(\text{SG}{=}0,\, \text{best.CPF},\, \text{best.BM},\, \text{attempts}{=}R \times A)$
\EndFunction
\end{algorithmic}
\end{algorithm}

In the reference implementation, the coding agent is instantiated as a single \gls{llm} with generic file and shell tools, while solver usage patterns, code templates, and repair heuristics are supplied through the skill files. More generally, however, the framework only assumes a coding agent that maps the task specification and current skill file to a candidate optimization program. This coding agent may therefore be realized by a single agent or by a multi-agent system without changing the evaluator or the outer skill-evolution loop. Its responsibility is limited to code generation and revision; execution, scoring, and stopping decisions remain outside the coding agent.

A deterministic evaluation pipeline, separate from the coding agent, executes each generated candidate program and evaluates the resulting design against fixed physical criteria. Both retry levels follow a generate, execute, feedback, and revise pattern \cite{openai_harness_2025}. Within one restart, attempts revise the same candidate program while retaining local contextual feedback, and the highest-ranked evaluated candidate from that restart is retained. Across restarts, the coding-agent session is reset, and only the best prior candidate together with a compact summary is carried forward. Evaluated candidates are ranked first by \gls{cpf} and then \gls{bm}, with earlier candidates preferred under ties; if no candidate reaches physical evaluation in any restart, the task is recorded as an execution failure. The procedure terminates as soon as any attempt reaches \gls{sg}$=1$. Because evaluation is performed outside the coding agent by a fixed procedure, the evaluation signal cannot be modified by the model itself.

\subsection{Problem Formulation and Evaluation Metrics}
\label{subsec:problem-formulation}

Each design task specifies a target optical response $\mathbf{y}^* \in \mathbb{R}^d$ (e.g., a transmission or reflection spectrum), from which a set of $K_i$ task-specific physical criteria $\{g_{i,j}\}_{j=1}^{K_i}$ is derived.
The objective is to find design parameters $\mathbf{p}_i \in \Omega_i$ such that the electromagnetic response computed by the forward solver $\mathcal{F}$ satisfies all criteria simultaneously:
\begin{equation}
    g_{i,j}\!\bigl(\mathcal{F}(\mathbf{p}_i)\bigr) \geq 0, \quad j = 1, \ldots, K_i,
    \label{eq:criteria}
\end{equation}
where $\mathcal{F}: \Omega_i \to \mathbb{R}^d$ maps design parameters to the simulated optical response using the {\solver} backend, and $g_{i,j}\!\bigl(\mathcal{F}(\mathbf{p}_i)\bigr)$ denotes the scalar margin for criterion $j$ of task $i$. For inequality criteria, the margin is the signed distance to the threshold; for target-matching criteria, it is the tolerance minus the deviation from the target. A nonnegative margin indicates that the criterion is satisfied, whereas a negative margin indicates violation.

We report five dataset-level metrics for evaluating design performance. \textbf{\Gls{se_rate}} measures the fraction of tasks for which the generated program executes successfully and the solver returns a valid simulation result:
\begin{equation}
    \text{SE} = \frac{1}{N}\sum_{i=1}^{N} \llbracket\,\text{code}_i\ \text{executes} \;\wedge\; \text{solver}_i\ \text{completes}\,\rrbracket.
    \label{eq:se}
\end{equation}
\textbf{\Gls{sg}} is the primary metric. For each task, we define a binary success indicator
\begin{equation}
    \text{SG}_i = \llbracket\,\forall\, j \in \{1,\ldots,K_i\}:\ g_{i,j}\!\bigl(\mathcal{F}(\mathbf{p}_i)\bigr) \geq 0\,\rrbracket,
    \label{eq:sg-task}
\end{equation}
and report the dataset-level success rate as
\begin{equation}
    \text{SG} = \frac{1}{N}\sum_{i=1}^{N} \text{SG}_i.
    \label{eq:sg}
\end{equation}
By definition, $\text{SG} \leq \text{SE}$; the gap $\text{SE} - \text{SG}$ captures tasks that execute successfully but fail to meet the physics requirements. \textbf{\Gls{cpf}} provides a finer-grained view by measuring the fraction of satisfied criteria for each task:
\begin{equation}
    \text{CPF}_i = \frac{1}{K_i}\sum_{j=1}^{K_i} \llbracket\,g_{i,j}\!\bigl(\mathcal{F}(\mathbf{p}_i)\bigr) \geq 0\,\rrbracket,
    \label{eq:cpf-task}
\end{equation}
and report the dataset-level average as
\begin{equation}
    \text{CPF} = \frac{1}{N}\sum_{i=1}^{N} \text{CPF}_i.
    \label{eq:cpf}
\end{equation}
\textbf{\Gls{bm}} quantifies the normalized margin of the least satisfied (bottleneck) criterion. For each task~$i$, we normalize the raw criterion margin by a criterion-specific scale,
\begin{equation}
    \hat{m}_{i,j} = \frac{g_{i,j}\!\bigl(\mathcal{F}(\mathbf{p}_i)\bigr)}{c_{i,j}},
    \qquad
    c_{i,j} =
    \begin{cases}
        \tau_{i,j}, & \text{for target-matching criteria},\\
        |t_{i,j}|, & \text{for inequality criteria with } t_{i,j} \neq 0,\\
        1, & \text{otherwise},
    \end{cases}
    \label{eq:norm-margin}
\end{equation}
where $t_{i,j}$ and $\tau_{i,j}$ denote the target and tolerance of criterion $(i,j)$, respectively. We then define the task-level bottleneck margin as
\begin{equation}
    \text{BM}_i = \min_{j}\, \hat{m}_{i,j},
    \label{eq:bm-task}
\end{equation}
and report the dataset-level average as
\begin{equation}
    \text{BM} = \frac{1}{N}\sum_{i=1}^{N} \text{BM}_i.
    \label{eq:bm}
\end{equation}
Equation~\ref{eq:bm-task} applies to tasks that the solver evaluates successfully; a task that fails to execute produces no criterion margins and is assigned $\text{BM}_i = 0$ in the average of Equation~\ref{eq:bm}, so \gls{bm} is best read alongside \gls{se_rate} rather than on its own. A larger \gls{bm} is otherwise better: a positive value indicates that all criteria are met with margin to spare, whereas a negative value reveals the magnitude of the worst violation, so a less negative \gls{bm} corresponds to a smaller worst-case violation. Finally, \textbf{Attempts} denotes the mean number of candidate programs generated per task, counting every generation across the restart and inner-retry loops of Algorithm~\ref{alg:mce-inverse-design} (at most $R\times A$); a task solved on the first generation counts as one.

\section{Experiments}
\label{sec:agentic-experiments}

\subsection{Experimental Setup}

We evaluate the framework with three settings that probe different kinds of generalization. The same-type setting asks whether evolved skills help on tasks drawn from template families represented during training. The two new-type settings, \NewTypeA and \NewTypeB, ask whether the same guidance helps on template families held out of the training set, and they reverse which families are held out: \NewTypeA trains on one family group and tests on the other; \NewTypeB does the reverse. Evaluating both directions tests whether a result reflects genuine transfer rather than the particular difficulty of a single task group.

All benchmark tasks are derived from six primary metasurface design template families (G1--G6) together with one auxiliary family ($G_{\mathrm{aux}}$) that balances the splits and serves as a full held-out family in the new-type settings (a test family in \NewTypeA, a training family in \NewTypeB). Each family is a group of closely related task templates that share physical structure and optimization objectives, and each template generates individual tasks that differ only in their numerical targets. These families have known feasible solution structures, ensuring that each instance corresponds to a physically meaningful inverse-design problem rather than an unconstrained code-generation exercise. We deliberately use parameterized design families with attainable optima rather than open-ended free-form topology optimization, so that the measured success rate reflects the framework's capability rather than the solvability of the underlying physics problem, providing a controlled setting for evaluating the framework itself. Each task specifies a target optical response together with a set of physical criteria, as defined in Section~\ref{subsec:problem-formulation}. Every split uses fixed 50/15/50 train/validation/test partitions. The same-type split keeps train, validation, and test within closely related template families, whereas the new-type splits hold out whole families: \NewTypeA trains on G1--G3, validates on G4, and tests on G5, G6, and $G_{\mathrm{aux}}$, while \NewTypeB is the same split with the train and test groups swapped and G4 still used for validation. Because each template is assigned to a single split, no task is used for both training and testing, and the \NewTypeA/\NewTypeB swap preserves this. During skill evolution, each iteration processes all 50 training tasks in three sub-batches of 20, 20, and 10, while the full validation split of 15 tasks is used for skill-file selection and the independent 50-task test split is reserved for final reporting.

The two new-type splits isolate different forms of generalization. In \NewTypeA, test families belong to device classes absent from both training and validation, so it measures whether a skill transfers to an unseen class. In \NewTypeB, the test families are reflective dielectric gratings, the same class as the validation family G4; because validation guides the skill-evolution loop, \NewTypeB instead measures whether the loop can autonomously acquire the conventions of a class it sees only through validation and apply them to held-out task sub-types it never trained on. A split-by-split summary of partitions is given in Appendix~\ref{app:agentic-dataset-splits}, and representative task specifications for each template family are provided in Appendix~\ref{app:agentic-task-examples}.

In this work, both the meta-agent and the coding agent use Claude Sonnet 4.6, and skill evolution is performed for four iterations. In each iteration, the coding agent applies the current skill file to the training tasks, and the meta-agent revises the skill file from the resulting per-task feedback. For each setting, we evaluate the baseline agent initialized with a starter skill file on the 50-task test split before skill evolution, and re-evaluate the selected post-training skill file on the same test split afterward, so the training and validation splits are used only for skill refinement and skill file selection, while both baseline and post-training results are measured on the same test dataset. Iteration-wise training and validation dynamics are summarized in Appendix~\ref{app:agentic-training-dynamics}.

\subsection{Main Results}

\begin{table}[htbp]
    \centering
    \caption{Test set performance: baseline (starter skill file) vs.\ post-train (evolved skills).}
    \label{tab:main-results}
    \begin{tabular}{lrrrrrr}
        \toprule
        Dataset & Condition & SG & SE & CPF & BM & Attempts \\
        \midrule
        Same-type & Starter-skill baseline & 38.0\% & 86.0\% & 0.510 & -2.380 & 4.10 \\
        Same-type & Post-train & 74.0\% & 100.0\% & 0.870 & -2.281 & 2.30 \\
        \rowcolor{yellow!15}
        Same-type & \textbf{Delta} & \textbf{+36.0pp} & \textbf{+14.0pp} & \textbf{+0.360} & \textbf{+0.099} & \textbf{-1.80} \\
        New-type-A & Starter-skill baseline & 92.0\% & 100.0\% & 0.960 & -4.626 & 1.60 \\
        New-type-A & Post-train & 90.0\% & 100.0\% & 0.950 & -2.092 & 1.68 \\
        \rowcolor{yellow!15}
        New-type-A & \textbf{Delta} & \textbf{-2.0pp} & \textbf{+0.0pp} & \textbf{-0.010} & \textbf{+2.534} & \textbf{+0.08} \\
        New-type-B & Starter-skill baseline & 20.0\% & 60.0\% & 0.340 & -0.148 & 4.36 \\
        New-type-B & Post-train & 90.0\% & 96.0\% & 0.930 & 0.140 & 1.78 \\
        \rowcolor{yellow!15}
        New-type-B & \textbf{Delta} & \textbf{+70.0pp} & \textbf{+36.0pp} & \textbf{+0.590} & \textbf{+0.288} & \textbf{-2.58} \\
        \bottomrule
    \end{tabular}
\end{table}

Table~\ref{tab:main-results} summarizes the main test-set results before and after skill evolution for the three settings, with \gls{sg} as the primary metric and \gls{se_rate}, \gls{cpf}, \gls{bm}, and Attempts as supporting metrics. Skill evolution improves the \SameType setting clearly, while the two new-type settings differ from each other: \NewTypeA starts near ceiling and is held there, whereas \NewTypeB starts low and improves sharply.

On \SameType tasks, skill evolution improves both success and efficiency: \gls{sg} increases from 38\% to 74\%, \gls{cpf} rises from 0.510 to 0.870, and mean attempts fall from 4.10 to 2.30. The two new-type splits then probe transfer in opposite directions. On \NewTypeA, the starter skill is already near ceiling (\gls{sg} = 92\%, 1.60 attempts), and the evolved skills hold success at 90\%; the mean \gls{bm} climbs from -4.626 to -2.092, though this gain comes almost entirely from fixing one severely violating case rather than from uniformly tighter margins. On \NewTypeB the held-out set is the harder family group, where the starter skill solves only 20\% of tasks; the gain is large across every metric: \gls{sg} 20\% to 90\%, \gls{se_rate} 0.60 to 0.96, \gls{cpf} 0.340 to 0.930, \gls{bm} -0.148 to +0.140, and attempts 4.36 to 1.78. These two directions mean different things: the \NewTypeB gain comes from the loop absorbing the held-out class's convention through validation, whereas \NewTypeA stays near ceiling because its starter skill already covers that class, not because evolution adds much.

\subsection{Outcomes}

\begin{figure}[htb!]
    \centering
    \includegraphics[width=\textwidth]{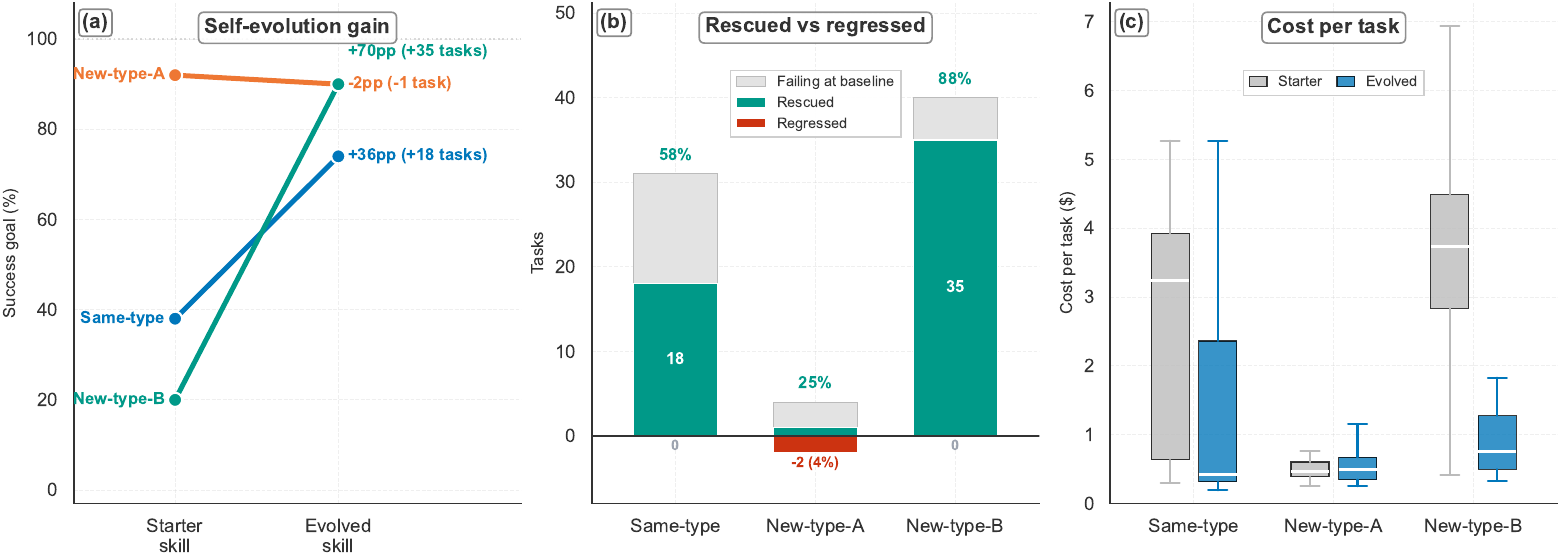}
    \caption{Outcomes of skill evolution across the three settings (\SameType, \NewTypeA, \NewTypeB), on held-out test tasks. (a) Self-evolution gain: each line traces \gls{sg} from the starter skill to the evolved skill, with signed delta labels in percentage points. (b) Rescued vs. regressed: per setting, the number of tasks moving from failing to passing (teal, rescued) or passing to failing (red, regressed); the faded grey pool is the baseline-failing denominator. (c) Cost per task: box plots of total \gls{llm} spend per task (USD) for the starter and evolved skill in each setting.}
    \label{fig:outcomes}
\end{figure}

Figure~\ref{fig:outcomes} presents these results from three complementary views: the self-evolution gain as a slope from starter to evolved skill, the balance of rescued and regressed tasks, and cost per task. The rescued-versus-regressed view confirms that the gains do not come at the expense of tasks that already pass: on \SameType, 18 of the 31 baseline failures are rescued (58\%) with zero regressions, and on \NewTypeB, 35 of 40 are rescued (88\%) with zero regressions. \NewTypeA, already near ceiling, rescues one of its four baseline failures but regresses two previously passing tasks, for a net change of one task (92\% to 90\%), so its success rate is not strictly monotone. The regression is an optimization near-miss rather than a breakage: \gls{se_rate} stays at 100\% and \gls{cpf} at 0.95, so every program still runs and returns a valid design, and the two regressed tasks fail only by falling short of a target criterion. Panel (c) shows that skill evolution also lowers cost where the starter skill was inefficient: the median cost per task falls from \$3.24 to \$0.42 on \SameType (about 87\%) and from \$3.73 to \$0.76 on \NewTypeB (about 80\%), while \NewTypeA, already inexpensive because its near-ceiling starter needs few attempts, stays low (\$0.46 to \$0.50). These reductions reflect a shift from broad exploratory search toward shorter revision cycles, detailed per tool in Figure~\ref{fig:si-tool-composition}.

\subsection{Mechanism of Skill Evolution}

\begin{figure}[htb!]
    \centering
    \includegraphics[width=\textwidth]{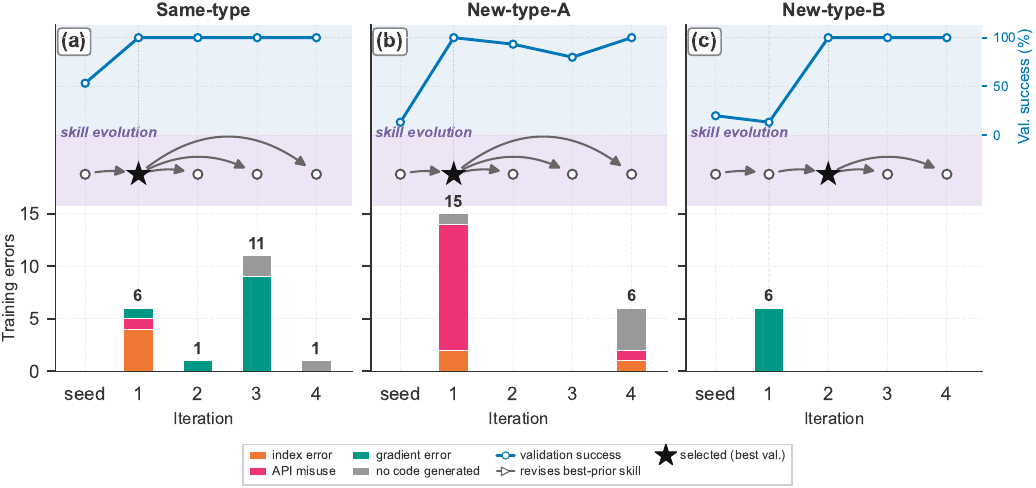}
    \caption{Mechanism of skill evolution, one panel per setting (\SameType, \NewTypeA, \NewTypeB). Within each panel, the blue line is the validation success rate (\gls{sg} on the held-out validation family, right axis) across skill-evolution iterations; the \emph{skill evolution} band shows the elitist search, where each node is a candidate skill file, an arrow marks a revision of the current best skill, and the star marks the iteration whose skill stays best on validation and becomes the final evolved skill; the stacked bars are the training-time error composition per iteration (left axis), with bar totals annotated. Because the loop keeps the best-on-validation skill and discards regressions, a later high-error candidate is a rejected exploration step, not a regression in the final skill. Infrastructure failures such as connection errors and timeouts are excluded. For \NewTypeB the seed marks the zero-shot baseline on the test families (its iteration~0 was scored there), whereas iterations~1 to~4 are on the validation family.}
    \label{fig:mechanism}
\end{figure}

Figure~\ref{fig:mechanism} traces how each setting arrives at its final skill under the elitist selection of Section~\ref{subsec:framework-overview}: the star marks the candidate that stays best on the validation family and is carried forward as the final evolved skill. On \SameType (panel a), validation success rises from 53\% at the seed skill to 100\% at iteration~1. No later candidate surpasses it: iterations~2 to~4 revise iteration~1 but only match its validation success, so iteration~1 is never replaced and remains the final evolved skill. The error bars show the code-level errors encountered on the training tasks at each iteration: index errors appear only in the first iteration, whereas the later revision attempts mostly resurface gradient errors, peaking at iteration~3 with eleven errors that are almost all gradient-related. Because iteration~1 already reaches the 100\% validation-success ceiling, these later candidates can at best tie it and never score higher, so the elitist rule keeps iteration~1 as the incumbent and carries it through to the end; the iteration~3 spike reflects new optimization patterns that break the computational graph, exactly the kind of regression the elitist rule is there to discard.

On \NewTypeA (panel b), the starter skill already solves 92\% of the 50 held-out test tasks, yet reaches only 13\% on the 15-task validation family that drives the loop. Iteration~1 lifts validation success to 100\% and becomes the skill carried forward; iterations~2 and~3 fall back to 93\% and 80\%, and although iteration~4 again reaches 100\% it does not surpass iteration~1 and is not selected. The first iteration's training errors are dominated by API-misuse, which then largely disappears. Because the test families were already near ceiling, the refinement that lifted validation success is not what they were missing, so the evolved skill stays close to the starter on the test set. This matches the near-unchanged test success and the 100\% execution rate in Figure~\ref{fig:outcomes}: with valid, running programs already in hand, the residual test misses are optimization near-misses rather than failures the validation signal could repair.

On \NewTypeB (panel c), validation success is only 13\% at iteration~1 and then jumps to 100\% at iteration~2; that candidate is the first to reach ceiling and is not surpassed, so it is the final evolved skill. This single step is where the loop acquires the family-group convention it was initially missing, the change behind the headline rise from 20\% to 90\% test success. Because the validation family is itself in the test families' device class, this is within-class learning from validation rather than transfer to an unseen class; \NewTypeA, by contrast, stays near ceiling with or without evolution. To make the learned changes directly inspectable, Appendix~\ref{app:agentic-skill-case-study} compares the starter and evolved skill files for the \SameType run side by side, showing how the skill grows from a minimal solver-setup guide into an operational protocol that reads prior results, turns recurring failures into explicit rules, and selects strategies by task type.

\section{Discussion and Conclusion}
\label{sec:agentic-discussion}

Overall, this paper argues that a central bottleneck in metasurface inverse design is not the absence of a physics solver, but the difficulty of repeatedly turning task specifications into working optimization code. The proposed self-evolving framework addresses that bottleneck by keeping the coding agent, solver, and evaluator fixed while evolving explicit skill files from solver-grounded per-task feedback. In this formulation, the adapted object is the workflow guidance itself rather than the model weights.

How much that guidance helps, and how far it carries beyond the families it was shaped on, is what the three evaluation settings measure. On \SameType, skill evolution improves success and efficiency together, raising \gls{sg} from 38\% to 74\%. The two new-type settings then probe generalization to families held out of training. Both withhold the test families' device class from training, so both reach beyond the trained families; they differ only in whether that class is visible during validation, the signal that drives skill evolution. In \NewTypeA, the held-out families (G5 plasmonic, G6 PbTe transmissive, and $G_{\mathrm{aux}}$ TiO$_2$ transmissive) are absent from both training and validation; the starter skill already sits near ceiling (\gls{sg} = 92\%), and the evolved skill holds success at 90\%, a net change of one task (one rescued, two regressed), while the mean \gls{bm} rises from $-4.626$ to $-2.092$, a shift dominated by a single outlier rather than a broad improvement. Because that skill never encountered those families, its sustained 90\% indicates that evolution preserves the starter's existing cross-class competence rather than creating it. In \NewTypeB, the test families are reflective dielectric gratings, the same class as the validation family G4, which the starter skill handles at only 20\%; here skill evolution raises held-out \gls{sg} to 90\% by autonomously acquiring the missing class convention from validation feedback and generalizing it to held-out task structures it never trained on. Taken together, the two settings illuminate different facets of the method: \NewTypeB shows the loop picking up a missing convention from validation alone and extending it to untrained task structures (+70pp), whereas \NewTypeA shows that adapting on unrelated families leaves the generalization the starter already had untouched. Broad cross-class binary generalization is not established by a single pair of swapped splits, but both directions show that the evolved skill file is not specific to the families it was developed on.

The evolved skill files already generalize beyond their training families; the framework that produces them is general by construction. The skill-evolution loop assumes only a coding agent that turns a task specification and the current skill file into a candidate optimization program, a deterministic evaluator that scores each result by physical simulation, and a human-readable skill file that accumulates reusable workflow knowledge. The base model, solver, and evaluator are interchangeable modules, and nothing in the loop encodes metasurface physics or {\solver} internals, so the same loop should carry over to other agentic scientific computing and inverse design problems by swapping in a different solver, evaluator, and starter skill. In this view, \glspl{llm} are most useful as programmable interfaces around reliable computational engines rather than as replacements for the solver itself, and skill evolution is the mechanism that turns solver-grounded feedback into explicit, inspectable, and reusable workflow knowledge. Being plain text, each skill file can be read, audited, and extended directly. The same file can seed a new problem without any retraining, which lowers the combined computational-electromagnetics and software-engineering barrier that still limits accessibility and cross-disciplinary use. The present evidence remains bounded by one solver stack, one benchmark design, and one base model; the natural next steps are to instantiate the same loop on additional solver backends and problem domains, and to test harder cross-regime and multiphysics transfer on larger, more diverse benchmarks.

\section*{Data Availability}

The code of the proposed framework will be released at \url{https://github.com/yi-huang-1/evo-metaoptics} upon publication.

\bibliographystyle{IEEEtran}
\bibliography{agent_inverse_design}

@article{chenReviewMetasurfacesPhysics2016,
  title = {A Review of Metasurfaces: Physics and Applications},
  shorttitle = {A Review of Metasurfaces},
  author = {Chen, Hou-Tong and Taylor, Antoinette J and Yu, Nanfang},
  year = 2016,
  month = jul,
  journal = {Reports on Progress in Physics},
  volume = {79},
  number = {7},
  pages = {076401},
  issn = {0034-4885, 1361-6633},
  doi = {10.1088/0034-4885/79/7/076401},
  urldate = {2025-04-19},
  langid = {english}
}

@article{dongAchromaticSingleMetalens2024,
  title = {Achromatic {{Single Metalens Imaging}} via {{Deep Neural Network}}},
  author = {Dong, Yunxi and Zheng, Bowen and Li, Hang and Tang, Hong and Zhao, Huan and Huang, Yi and An, Sensong and Zhang, Hualiang},
  year = 2024,
  month = apr,
  journal = {ACS Photonics},
  volume = {11},
  number = {4},
  pages = {1645--1656},
  issn = {2330-4022, 2330-4022},
  doi = {10.1021/acsphotonics.3c01870},
  urldate = {2024-12-02},
  copyright = {https://doi.org/10.15223/policy-029},
  langid = {english}
}

@article{dongAdvancedDeepLearning2025,
  title = {Advanced Deep Learning Approaches in Metasurface Modeling and Design: {{A}} Review},
  shorttitle = {Advanced Deep Learning Approaches in Metasurface Modeling and Design},
  author = {Dong, Yunxi and An, Sensong and Jiang, Haoyue and Zheng, Bowen and Tang, Hong and Huang, Yi and Zhao, Huan and Zhang, Hualiang},
  year = 2025,
  month = jan,
  journal = {Progress in Quantum Electronics},
  volume = {99},
  pages = {100554},
  issn = {00796727},
  doi = {10.1016/j.pquantelec.2025.100554},
  urldate = {2026-03-31},
  langid = {english}
}

@misc{fangComprehensiveSurveySelfEvolving2025,
  title = {A {{Comprehensive Survey}} of {{Self-Evolving AI Agents}}: {{A New Paradigm Bridging Foundation Models}} and {{Lifelong Agentic Systems}}},
  shorttitle = {A {{Comprehensive Survey}} of {{Self-Evolving AI Agents}}},
  author = {Fang, Jinyuan and Peng, Yanwen and Zhang, Xi and Wang, Yingxu and Yi, Xinhao and Zhang, Guibin and Xu, Yi and Wu, Bin and Liu, Siwei and Li, Zihao and Ren, Zhaochun and Aletras, Nikos and Wang, Xi and Zhou, Han and Meng, Zaiqiao},
  year = 2025,
  month = aug,
  number = {arXiv:2508.07407},
  eprint = {2508.07407},
  primaryclass = {cs},
  publisher = {arXiv},
  doi = {10.48550/arXiv.2508.07407},
  urldate = {2025-12-09},
  archiveprefix = {arXiv},
  langid = {english}
}

@article{gaoInverseDesignFlat2022,
  title = {Inverse Design in Flat Optics},
  author = {Gao, Yubin and Chen, Qikai and Pian, Sijie and Ma, Yaoguang},
  year = 2022,
  month = dec,
  journal = {Photonics and Nanostructures - Fundamentals and Applications},
  volume = {52},
  pages = {101074},
  issn = {1569-4410},
  doi = {10.1016/j.photonics.2022.101074},
  urldate = {2022-12-24},
  langid = {english}
}

@misc{gaoSurveySelfEvolvingAgents2025,
  title = {A {{Survey}} of {{Self-Evolving Agents}}: {{On Path}} to {{Artificial Super Intelligence}}},
  shorttitle = {A {{Survey}} of {{Self-Evolving Agents}}},
  author = {Gao, Huan-ang and Geng, Jiayi and Hua, Wenyue and Hu, Mengkang and Juan, Xinzhe and Liu, Hongzhang and Liu, Shilong and Qiu, Jiahao and Qi, Xuan and Wu, Yiran and Wang, Hongru and Xiao, Han and Zhou, Yuhang and Zhang, Shaokun and Zhang, Jiayi and Xiang, Jinyu and Fang, Yixiong and Zhao, Qiwen and Liu, Dongrui and Ren, Qihan and Qian, Cheng and Wang, Zhenhailong and Hu, Minda and Wang, Huazheng and Wu, Qingyun and Ji, Heng and Wang, Mengdi},
  year = 2025,
  month = aug,
  number = {arXiv:2507.21046},
  eprint = {2507.21046},
  primaryclass = {cs},
  publisher = {arXiv},
  doi = {10.48550/arXiv.2507.21046},
  urldate = {2025-12-13},
  archiveprefix = {arXiv},
  langid = {english}
}

@misc{heEvoTestEvolutionaryTestTime2026,
  title = {{{EvoTest}}: {{Evolutionary Test-Time Learning}} for {{Self-Improving Agentic Systems}}},
  shorttitle = {{{EvoTest}}},
  author = {He, Yufei and Liu, Juncheng and Liu, Yue and Li, Yibo and Cao, Tri and Hu, Zhiyuan and Xu, Xinxing and Hooi, Bryan},
  year = 2026,
  month = apr,
  number = {arXiv:2510.13220},
  eprint = {2510.13220},
  primaryclass = {cs.AI},
  publisher = {arXiv},
  doi = {10.48550/arXiv.2510.13220},
  urldate = {2026-06-22},
  archiveprefix = {arXiv}
}

@inproceedings{huang2024differentiable,
  title = {Differentiable Inverse Design of Free-Form Meta-Optics Using Multiplicative Filter Network},
  booktitle = {2024 International Applied Computational Electromagnetics Society Symposium ({{ACES}})},
  author = {Huang, Yi and Dong, Yunxi and Zhao, Huan and Tang, Hong and Zheng, Bowen and Zhang, Hualiang},
  year = 2024,
  pages = {1--2},
  publisher = {IEEE}
}

@inproceedings{huang3DPrintedMillimeterWaveFreeForm2024,
  title = {A {{3D-Printed Millimeter-Wave Free-Form Metasurface Based}} on {{Automatic Differentiable Inverse Design}}},
  booktitle = {2024 {{IEEE}}/{{MTT-S International Microwave Symposium}} - {{IMS}} 2024},
  author = {Huang, Yi and Tang, Hong and Zhao, Huan and Dong, Yunxi and Zheng, Bowen and Zhang, Hualiang},
  year = 2024,
  month = jun,
  pages = {559--562},
  publisher = {IEEE},
  address = {Washington, DC, USA},
  doi = {10.1109/IMS40175.2024.10600252},
  urldate = {2024-12-02},
  copyright = {https://doi.org/10.15223/policy-029},
  isbn = {979-8-3503-7504-6},
  langid = {english}
}

@article{huangEigendecompositionfreeInverseDesign2024,
  title = {Eigendecomposition-Free Inverse Design of Meta-Optics Devices},
  author = {Huang, Yi and Zhu, Ziwei and Dong, Yunxi and Tang, Hong and Zheng, Bowen and Podolskiy, Viktor A. and Zhang, Hualiang},
  year = 2024,
  month = apr,
  journal = {Optics Express},
  volume = {32},
  number = {8},
  pages = {13986--13997},
  publisher = {Optica Publishing Group},
  doi = {10.1364/OE.514347}
}

@inproceedings{huangInverseDesignPhotonic2023,
  title = {Inverse {{Design}} of {{Photonic Structures Using Automatic Differentiable Rigorous Diffraction Interface Theory}}},
  booktitle = {{{CLEO}} 2023},
  author = {Huang, Yi and Tang, Hong and Zheng, Bowen and Dong, Yunxi and Haerinia, Mohammad and Podolskiy, Viktor A. and Zhang, Hualiang},
  year = 2023,
  month = may,
  pages = {JTu2A.119},
  publisher = {Optica Publishing Group},
  doi = {10.1364/CLEO_AT.2023.JTu2A.119},
  urldate = {2023-10-12},
  copyright = {\&\#169; 2023 The Author(s)},
  langid = {english}
}

@article{huangMCPenabledLLMMetaoptics2025,
  title = {{{MCP-enabled LLM}} for Meta-Optics Inverse Design: Leveraging Differentiable Solver without {{LLM}} Expertise},
  shorttitle = {{{MCP-enabled LLM}} for Meta-Optics Inverse Design},
  author = {Huang, Yi and Zheng, Bowen and Dong, Yunxi and Tang, Hong and Zhao, Huan and Shawon, Rakibul Hasan and An, Sensong and Zhang, Hualiang},
  year = 2025,
  month = dec,
  journal = {Nanophotonics},
  issn = {2192-8606, 2192-8614},
  doi = {10.1515/nanoph-2025-0507},
  urldate = {2025-12-09},
  copyright = {http://creativecommons.org/licenses/by/4.0},
  langid = {english}
}

@article{jensenTopologyOptimizationNanophotonics2011,
  title = {Topology Optimization for Nano-Photonics},
  author = {Jensen, J.s. and Sigmund, O.},
  year = 2011,
  journal = {Laser \& Photonics Reviews},
  volume = {5},
  number = {2},
  pages = {308--321},
  issn = {1863-8899},
  doi = {10.1002/lpor.201000014},
  urldate = {2022-12-25},
  langid = {english}
}

@article{kimNanophotonicDeviceDesign2025,
  title = {Nanophotonic Device Design Based on Large Language Models: Multilayer and Metasurface Examples},
  shorttitle = {Nanophotonic Device Design Based on Large Language Models},
  author = {Kim, Myungjoon and Park, Hyeonjin and Shin, Jonghwa},
  year = 2025,
  month = apr,
  journal = {Nanophotonics},
  volume = {14},
  number = {8},
  pages = {1273--1282},
  issn = {2192-8614},
  doi = {10.1515/nanoph-2024-0674},
  urldate = {2025-04-25},
  copyright = {http://creativecommons.org/licenses/by/4.0},
  langid = {english}
}

@article{liEmpoweringMetasurfacesInverse2022,
  title = {Empowering {{Metasurfaces}} with {{Inverse Design}}: {{Principles}} and {{Applications}}},
  shorttitle = {Empowering {{Metasurfaces}} with {{Inverse Design}}},
  author = {Li, Zhaoyi and Pestourie, Rapha{\"e}l and Lin, Zin and Johnson, Steven G. and Capasso, Federico},
  year = 2022,
  month = jul,
  journal = {ACS Photonics},
  volume = {9},
  number = {7},
  pages = {2178--2192},
  publisher = {American Chemical Society},
  doi = {10.1021/acsphotonics.1c01850},
  urldate = {2022-12-24}
}

@article{luAgenticFrameworkAutonomous2025,
  title = {An {{Agentic Framework}} for {{Autonomous Metamaterial Modeling}} and {{Inverse Design}}},
  author = {Lu, Darui and Malof, Jordan M. and Padilla, Willie J.},
  year = 2025,
  month = nov,
  journal = {ACS Photonics},
  volume = {12},
  number = {11},
  pages = {6071--6080},
  issn = {2330-4022, 2330-4022},
  doi = {10.1021/acsphotonics.5c01514},
  urldate = {2026-03-15},
  copyright = {https://doi.org/10.15223/policy-029},
  langid = {english}
}

@misc{luLearningElectromagneticMetamaterial2025,
  title = {Learning {{Electromagnetic Metamaterial Physics With ChatGPT}}},
  author = {Lu, Darui and Deng, Yang and Malof, Jordan M. and Padilla, Willie J.},
  year = 2025,
  month = feb,
  number = {arXiv:2404.15458},
  eprint = {2404.15458},
  primaryclass = {physics},
  publisher = {arXiv},
  doi = {10.48550/arXiv.2404.15458},
  urldate = {2025-06-15},
  archiveprefix = {arXiv},
  langid = {english}
}

@misc{luoEmpiricalStudyCatastrophic2025,
  title = {An {{Empirical Study}} of {{Catastrophic Forgetting}} in {{Large Language Models During Continual Fine-tuning}}},
  author = {Luo, Yun and Yang, Zhen and Meng, Fandong and Li, Yafu and Zhou, Jie and Zhang, Yue},
  year = 2025,
  month = jan,
  number = {arXiv:2308.08747},
  eprint = {2308.08747},
  primaryclass = {cs},
  publisher = {arXiv},
  doi = {10.48550/arXiv.2308.08747},
  urldate = {2026-03-20},
  archiveprefix = {arXiv}
}

@article{lupoiuMultiagenticFrameworkRealtime2025a,
  title = {A Multi-Agentic Framework for Real-Time, Autonomous Freeform Metasurface Design},
  author = {Lupoiu, Robert and Shao, Yixuan and Dai, Tianxiang and Mao, Chenkai and Ed{\'e}e, Kofi and Fan, Jonathan A.},
  year = 2025,
  month = oct,
  journal = {Science Advances},
  volume = {11},
  number = {44},
  pages = {eadx8006},
  issn = {2375-2548},
  doi = {10.1126/sciadv.adx8006},
  urldate = {2026-03-15},
  langid = {english}
}

@article{maOptoGPTFoundationModel2024,
  title = {{{OptoGPT}}: {{A}} Foundation Model for Inverse Design in Optical Multilayer Thin Film Structures},
  shorttitle = {{{OptoGPT}}},
  author = {Ma, Taigao and Wang, Haozhu and Guo, L. Jay and {Department of Physics, University of Michigan, Ann Arbor, Michigan 48109, USA} and {Department of Electrical Engineering and Computer Science, University of Michigan, Ann Arbor, Michigan 48109, USA}},
  year = 2024,
  journal = {Opto-Electronic Advances},
  volume = {7},
  number = {7},
  pages = {240062--240062},
  issn = {2096-4579},
  doi = {10.29026/oea.2024.240062},
  urldate = {2025-06-30},
  langid = {english}
}

@article{moleskyInverseDesignNanophotonics2018,
  title = {Inverse Design in Nanophotonics},
  author = {Molesky, Sean and Lin, Zin and Piggott, Alexander Y. and Jin, Weiliang and Vuckovi{\'c}, Jelena and Rodriguez, Alejandro W.},
  year = 2018,
  month = nov,
  journal = {Nature Photonics},
  volume = {12},
  number = {11},
  pages = {659--670},
  publisher = {Nature Publishing Group},
  issn = {1749-4893},
  doi = {10.1038/s41566-018-0246-9},
  urldate = {2022-12-21},
  copyright = {2018 Springer Nature Limited},
  langid = {english}
}

@misc{openai_harness_2025,
  title = {Harness Engineering: Leveraging {{Codex}} in an Agent-First World},
  author = {{OpenAI}},
  year = 2026,
  month = feb,
  howpublished = {https://openai.com/index/harness-engineering/}
}

@article{ouAdvancesMetaOpticsMetasurfaces2023,
  title = {Advances in {{Meta-Optics}} and {{Metasurfaces}}: {{Fundamentals}} and {{Applications}}},
  shorttitle = {Advances in {{Meta-Optics}} and {{Metasurfaces}}},
  author = {Ou, Kai and Wan, Hengyi and Wang, Guangfeng and Zhu, Jingyuan and Dong, Siyu and He, Tao and Yang, Hui and Wei, Zeyong and Wang, Zhanshan and Cheng, Xinbin},
  year = 2023,
  month = jan,
  journal = {Nanomaterials},
  volume = {13},
  number = {7},
  pages = {1235},
  publisher = {Multidisciplinary Digital Publishing Institute},
  issn = {2079-4991},
  doi = {10.3390/nano13071235},
  urldate = {2023-12-03},
  copyright = {http://creativecommons.org/licenses/by/3.0/},
  langid = {english}
}

@article{parkFreeformOptimizationNanophotonic2022,
  title = {Free-Form Optimization of Nanophotonic Devices: From Classical Methods to Deep Learning},
  shorttitle = {Free-Form Optimization of Nanophotonic Devices},
  author = {Park, Juho and Kim, Sanmun and Nam, Daniel Wontae and Chung, Haejun and Park, Chan Y. and Jang, Min Seok},
  year = 2022,
  month = may,
  journal = {Nanophotonics},
  volume = {11},
  number = {9},
  pages = {1809--1845},
  issn = {2192-8614},
  doi = {10.1515/nanoph-2021-0713},
  urldate = {2026-03-13},
  copyright = {http://creativecommons.org/licenses/by/4.0},
  langid = {english}
}

@misc{schmid_agent_harness_2026,
  title = {The Importance of Agent Harness in 2026},
  author = {Schmid, Philipp},
  year = 2026,
  howpublished = {https://www.philschmid.de/agent-harness-2026}
}

@misc{yeMetaContextEngineering2026,
  title = {Meta {{Context Engineering}} via {{Agentic Skill Evolution}}},
  author = {Ye, Haoran and He, Xuning and Arak, Vincent and Dong, Haonan and Song, Guojie},
  year = 2026,
  month = feb,
  number = {arXiv:2601.21557},
  eprint = {2601.21557},
  primaryclass = {cs},
  publisher = {arXiv},
  doi = {10.48550/arXiv.2601.21557},
  urldate = {2026-02-17},
  archiveprefix = {arXiv},
  langid = {english}
}

@misc{zhangAgenticContextEngineering2025,
  title = {Agentic {{Context Engineering}}: {{Evolving Contexts}} for {{Self-Improving Language Models}}},
  shorttitle = {Agentic {{Context Engineering}}},
  author = {Zhang, Qizheng and Hu, Changran and Upasani, Shubhangi and Ma, Boyuan and Hong, Fenglu and Kamanuru, Vamsidhar and Rainton, Jay and Wu, Chen and Ji, Mengmeng and Li, Hanchen and Thakker, Urmish and Zou, James and Olukotun, Kunle},
  year = 2025,
  month = oct,
  number = {arXiv:2510.04618},
  eprint = {2510.04618},
  primaryclass = {cs},
  publisher = {arXiv},
  doi = {10.48550/arXiv.2510.04618},
  urldate = {2025-12-09},
  archiveprefix = {arXiv},
  langid = {english}
}

@article{zhangChatChipLarge2025,
  title = {Chat to Chip: Large Language Model Based Design of Arbitrarily Shaped Metasurfaces},
  shorttitle = {Chat to Chip},
  author = {Zhang, Huanshu and Kang, Lei and Campbell, Sawyer D. and Werner, Douglas H.},
  year = 2025,
  month = nov,
  journal = {Nanophotonics},
  volume = {14},
  number = {22},
  pages = {3625--3633},
  issn = {2192-8614},
  doi = {10.1515/nanoph-2025-0343},
  urldate = {2026-03-15},
  copyright = {http://creativecommons.org/licenses/by/4.0},
  langid = {english}
}

\newpage

\begin{appendices}

\nolinenumbers  
\setcounter{page}{1}  
\renewcommand{\thesection}{S\arabic{section}}
\setcounter{table}{0}  
\renewcommand{\thetable}{S\arabic{table}}  
\glsresetall

\numberwithin{equation}{section}                
\renewcommand{\theequation}{\thesection.\arabic{equation}} 
\numberwithin{figure}{section}
\renewcommand{\thefigure}{\thesection.\arabic{figure}} 

\section{Skill File Structure}
\label{app:agentic-skill-artifact}

The learned object studied in this work is a single, versioned skill file rather than an unstructured block of prompt text. Let \(S_{i,m}\) denote the active evolved skill after iteration \(i\) and sub-iteration \(m\). In the current implementation, this learned object is the Markdown file \texttt{iterN\_subM/.agents/skills/learning-context/SKILL.md}. The file is structured and validated against a minimal schema, including YAML frontmatter with \texttt{name} and \texttt{description} fields and a required \texttt{\#\# Skill Overview} section. The meta-agent directly rewrites this \texttt{SKILL.md}, so it is the principal learned object carried across the evolution loop.

For a per-task sample \(k\), the runtime assembles a sample-level package
\[
\mathcal{P}_{i,m,k} = \{ S_{i,m} \} \cup \mathcal{R},
\]
under \texttt{iterN\_subM/codegen/iter\_K/.agents/skills/learning-context/}, where \(\mathcal{R}\) is a deterministic \texttt{reference/} subtree. In the current runtime, this subtree contains static Markdown references for setup, materials and layers, patterning, sources and solving, phase and diffraction orders, amplitude and efficiency metrics, optimization, common pitfalls, and context usage. These reference documents are generated by the environment and are not themselves rewritten by the meta-agent during skill evolution.

After the final sub-iteration of iteration \(i\), the resulting skill is archived as
\[
\mathcal{A}_i = \operatorname{archive}(S_{i,m^{\star}}),
\]
stored at \texttt{meta\_agent/skills/learning-context-iterN/SKILL.md}. These archived snapshots form the historical skill memory consulted by later iterations and support analysis of the evolution history.

The observed directory structure is summarized in Listing~\ref{lst:agentic-skill-artifact-tree}. This representation makes explicit that the evolving core is a single \texttt{SKILL.md}, while execution-time packages additionally include a static \texttt{reference/} subtree, and completed iterations are preserved as archived snapshots.

\begin{lstlisting}[
    style=xml,
    caption={\textbf{Observed folder structure of the evolved skill file}},
    label={lst:agentic-skill-artifact-tree}
]
workspace/<run>/
|-- iterN_subM/
|   `-- .agents/skills/learning-context/
|       `-- SKILL.md
|           # versioned skill file S_{i,m} (the evolving object)
|           # directly rewritten by the meta-agent
|
|-- iterN_subM/codegen/iter_K/
|   `-- .agents/skills/learning-context/
|       |-- SKILL.md
|       |   # working copy of S_{i,m}
|       |   # used by one per-task sample, not independently evolvable
|       `-- reference/
|           |-- setup.md
|           |-- materials_and_layers.md
|           |-- patterning.md
|           |-- sources_and_solving.md
|           |-- phase_and_orders.md
|           |-- amplitude_and_efficiency.md
|           |-- optimization.md
|           |-- pitfalls.md
|           `-- context_usage.md
|               # static reference subtree R, generated by the runtime
|               # not evolvable within the current evolution loop
|
`-- meta_agent/skills/
    `-- learning-context-iterN/
        `-- SKILL.md
            # archived snapshot A_i copied from the evolved skill
            # preserved for history, not independently evolvable
\end{lstlisting}

Across sampled same-type and new-type runs, the evolved \texttt{SKILL.md} files consistently combine four kinds of guidance: methodology-level workflow rules, solver API and usage patterns, optimization and task-type-specific recipes, and failure-avoidance heuristics. Skill evolution therefore updates reusable textual guidance while keeping the base model, evaluator, physical solver, and held-out test split fixed. In this sense, this work studies the evolution of reusable, human-readable skill files rather than the adaptation of model weights.

\section{Dataset Split Summary}
\label{app:agentic-dataset-splits}

Rather than enumerating implementation-specific template identifiers, this appendix describes the benchmark used in this work in terms of abstract task families. A task instance is written as
\[
\tau = (\xi, \mathcal{X}, \mathcal{C}),
\]
where \(\xi\) denotes the fixed physical context, including wavelength, incidence condition, material setting, and lattice parameters, \(\mathcal{X}\) denotes the design space, and \(\mathcal{C}\) denotes the evaluation criteria. The benchmark is organized around six primary families \(G_1,\dots,G_6\), plus one auxiliary family \(G_{\mathrm{aux}}\), a non-primary template added to balance the fixed splits: it rounds out the Same-type validation partition and forms the third member of the held-out family group in the new-type settings, where it acts as a full test family (New-type-A) and a full training family (New-type-B). For transmissive phase-target tasks, the criteria take the form
\[
T(\lambda_0; x) \ge \tau_T,
\qquad
 d_{\mathbb{S}^1}\!\left(\phi(\lambda_0; x), \phi^{\star}\right) \le \varepsilon_{\phi},
\]
where \(d_{\mathbb{S}^1}\) denotes cyclic phase distance. Reflective tasks use one or two constraints on \(R\), evaluated across wavelength, incidence angle, or polarization depending on the family.

\begin{table}[htbp]
    \centering
    \caption{Abstract task-family taxonomy used to describe the benchmark in this work.}
    \label{tab:agentic-dataset-taxonomy}
    \footnotesize
    \setlength{\tabcolsep}{4pt}
    \begin{tabular}{p{2.5cm}p{4cm}p{5cm}p{5.5cm}}
        \toprule
        Family & Problem type & Design space & Criterion form \\
        \midrule
        \(G_1\) & Single-condition reflective design & Low-dimensional periodic reflective geometry & \(R(\lambda_0; x) \ge \tau_R\) \\
        \(G_2\) & Dual-angle reflective design & Low-dimensional periodic reflective geometry & \parbox[t]{5.5cm}{\raggedright \(\begin{aligned}[t]
R(\lambda_0, \theta_1; x) &\ge \tau_R, \\
R(\lambda_0, \theta_2; x) &\ge \tau_R
\end{aligned}\)} \\
        \(G_3\) & Polarization-selective reflective design & Low-dimensional periodic reflective geometry & \parbox[t]{5.5cm}{\raggedright \(\begin{aligned}[t]
R_{\mathrm{TE}}(\lambda_0; x) &\ge \tau_{\mathrm{hi}}, \\
R_{\mathrm{TM}}(\lambda_0; x) &\le \tau_{\mathrm{lo}}
\end{aligned}\)} \\
        \(G_4\) & Dual-wavelength reflective design & Low-dimensional periodic reflective geometry & \parbox[t]{5.5cm}{\raggedright \(\begin{aligned}[t]
R(\lambda_1; x) &\ge \tau_R, \\
R(\lambda_2; x) &\ge \tau_R
\end{aligned}\)} \\
        \(G_5\) & Plasmonic reflective design & Periodic reflective geometry in a plasmonic regime & \(R(\lambda_0; x) \ge \tau_R\) \\
        \(G_6\) & Transmissive phase-target metasurface design & Unit-cell geometry with family-dependent parameterization & \parbox[t]{5.5cm}{\raggedright \(\begin{aligned}[t]
&T(\lambda_0; x) \ge \tau_T, \\
&d_{\mathbb{S}^1}(\phi(\lambda_0; x), \phi^{\star}) \le \varepsilon_{\phi}
\end{aligned}\)} \\
        \(G_{\mathrm{aux}}\) & Additional transmissive phase-target template & Unit-cell geometry with a distinct visible-wavelength parameterization & Same criterion form as \(G_6\) \\
        \bottomrule
    \end{tabular}
\end{table}

The fixed 50/15/50 train, validation, and test partitions are defined at the level of sibling templates rather than by unconstrained random instance sampling. For a split with target size \(N\) and \(k\) constituent templates, each template contributes either \(\lfloor N/k \rfloor\) or \(\lceil N/k \rceil\) rows. This keeps the family mix balanced while preserving the intended same-type and family-heldout evaluation regimes.

\begin{table}[htbp]
    \centering
    \caption{Abstract composition of the finalized splits for the three evaluation settings.}
    \label{tab:agentic-dataset-splits}
    \small
    \begin{tabular}{lllp{5.6cm}p{4.0cm}}
        \toprule
        Setting & Split & Tasks & Family coverage & Interpretation \\
        \midrule
        Same-type & Train & 50 & One sibling template from each of \(G_1\) through \(G_6\) & Same-type training over all six primary families \\
        Same-type & Validation & 15 & Held-out sibling of \(G_1\), plus \(G_{\mathrm{aux}}\) & Mild generalization check using closely related sibling templates \\
        Same-type & Test & 50 & Held-out sibling templates from \(G_2\) through \(G_6\) & Same-type test on sibling templates not used during training \\
        \midrule
        New-type-A & Train & 50 & \(G_1\), \(G_2\), and \(G_3\) & Training restricted to single-condition, dual-angle, and polarization-selective reflection \\
        New-type-A & Validation & 15 & \(G_4\) & Validation on an unseen dual-wavelength family \\
        New-type-A & Test & 50 & \(G_5\), \(G_6\), and \(G_{\mathrm{aux}}\) & Family-heldout test across plasmonic and transmissive phase-target regimes \\
        New-type-B & Train & 50 & \(G_5\), \(G_6\), and \(G_{\mathrm{aux}}\) & Training on the plasmonic and transmissive families that New-type-A holds out \\
        New-type-B & Validation & 15 & \(G_4\) & Same dual-wavelength family used for New-type-A validation \\
        New-type-B & Test & 50 & \(G_1\), \(G_2\), and \(G_3\) & Family-heldout test on the reflective grating families used to train New-type-A \\
        \bottomrule
    \end{tabular}
\end{table}

Under \SameType, the splits stay close, drawing on sibling templates that share objective structure and physical regime and differ only in task instances. The two new-type settings instead use disjoint families across train, validation, and test, so transfer must cross both dimensions, which is why Section~\ref{sec:agentic-experiments} treats them as family-heldout. They are role-swaps sharing one held-out validation family, \(G_4\): \NewTypeA trains on the reflective gratings and tests on the plasmonic and transmissive families, while \NewTypeB reverses this. Crucially, \(G_4\) is itself a reflective grating, so it shares a device class with the test families of \NewTypeB but not with those of \NewTypeA; the main text therefore reads \NewTypeB as within-class acquisition and \NewTypeA as cross-class transfer.

\section{Representative Task Specifications}
\label{app:agentic-task-examples}

This appendix provides one representative task specification from each of the template families described in Section~\ref{app:agentic-dataset-splits}. Within each family, individual task instances are generated by sampling physical parameters (wavelength, material indices, periodicity, and criterion thresholds) from predefined ranges, so that tasks within the same family share structural form but differ in numerical specifications. Each task is presented in the JSON format consumed by the coding agent and the deterministic evaluator. The \texttt{query} field contains the natural-language task description provided to the coding agent, \texttt{gt\_eval} defines the evaluation criteria used by the deterministic evaluator, and \texttt{reference} records the DOI of the originating publication.

\begin{lstlisting}[style=json, caption={$G_1$: Single-condition reflective design.}, label={lst:task-g1}]
{
  (*@\textcolor{jsonkey}{"query"}@*): "Design a single-layer reflective grating at 0.632 um from RCWA-TF example settings. Use period 0.4777 um, high-index layer n=2.436, k=0.000, substrate n=1.363. Use TE polarization at normal incidence and maximize total reflection efficiency.",
  (*@\textcolor{jsonkey}{"gt\_eval"}@*): {
    (*@\textcolor{jsonkey}{"wavelength\_um"}@*): [0.632],
    (*@\textcolor{jsonkey}{"criteria"}@*): [
      {(*@\textcolor{jsonkey}{"metric"}@*): "total_reflection", (*@\textcolor{jsonkey}{"params"}@*): {(*@\textcolor{jsonkey}{"wavelength\_index"}@*): 0},
       (*@\textcolor{jsonkey}{"operation"}@*): ">=", (*@\textcolor{jsonkey}{"target"}@*): 0.8}
    ]
  },
  (*@\textcolor{jsonkey}{"reference"}@*): "10.1038/s42005-021-00568-6"
}
\end{lstlisting}

\begin{lstlisting}[style=json, caption={$G_2$: Dual-angle reflective design.}, label={lst:task-g2}]
{
  (*@\textcolor{jsonkey}{"query"}@*): "Design a single-layer reflective grating at 0.632 um for two angles (0 deg and 5.0 deg). Use period 0.4777 um, high-index layer n=2.436, substrate n=1.363, and TE polarization. Optimize geometry and layer thickness to maximize the product of reflectances at both angles.",
  (*@\textcolor{jsonkey}{"gt\_eval"}@*): {
    (*@\textcolor{jsonkey}{"wavelength\_um"}@*): [0.632],
    (*@\textcolor{jsonkey}{"criteria"}@*): [
      {(*@\textcolor{jsonkey}{"metric"}@*): "total_reflection", (*@\textcolor{jsonkey}{"params"}@*): {(*@\textcolor{jsonkey}{"source\_index"}@*): 0, (*@\textcolor{jsonkey}{"wavelength\_index"}@*): 0},
       (*@\textcolor{jsonkey}{"operation"}@*): ">=", (*@\textcolor{jsonkey}{"target"}@*): 0.7},
      {(*@\textcolor{jsonkey}{"metric"}@*): "total_reflection", (*@\textcolor{jsonkey}{"params"}@*): {(*@\textcolor{jsonkey}{"source\_index"}@*): 1, (*@\textcolor{jsonkey}{"wavelength\_index"}@*): 0},
       (*@\textcolor{jsonkey}{"operation"}@*): ">=", (*@\textcolor{jsonkey}{"target"}@*): 0.7}
    ]
  },
  (*@\textcolor{jsonkey}{"reference"}@*): "10.1038/s42005-021-00568-6"
}
\end{lstlisting}

\begin{lstlisting}[style=json, caption={$G_3$: Polarization-selective reflective design.}, label={lst:task-g3}]
{
  (*@\textcolor{jsonkey}{"query"}@*): "Design a reflective line grating at 0.632 um with period 0.5118 um. Use high-index layer n=2.436, substrate n=1.363. Optimize width and thickness so TE reflection is high while TM reflection is low at normal incidence.",
  (*@\textcolor{jsonkey}{"gt\_eval"}@*): {
    (*@\textcolor{jsonkey}{"wavelength\_um"}@*): [0.632],
    (*@\textcolor{jsonkey}{"criteria"}@*): [
      {(*@\textcolor{jsonkey}{"metric"}@*): "total_reflection", (*@\textcolor{jsonkey}{"params"}@*): {(*@\textcolor{jsonkey}{"source\_index"}@*): 0, (*@\textcolor{jsonkey}{"wavelength\_index"}@*): 0},
       (*@\textcolor{jsonkey}{"operation"}@*): ">=", (*@\textcolor{jsonkey}{"target"}@*): 0.8},
      {(*@\textcolor{jsonkey}{"metric"}@*): "total_reflection", (*@\textcolor{jsonkey}{"params"}@*): {(*@\textcolor{jsonkey}{"source\_index"}@*): 1, (*@\textcolor{jsonkey}{"wavelength\_index"}@*): 0},
       (*@\textcolor{jsonkey}{"operation"}@*): "<=", (*@\textcolor{jsonkey}{"target"}@*): 0.2}
    ]
  },
  (*@\textcolor{jsonkey}{"reference"}@*): "10.1038/s42005-021-00568-6"
}
\end{lstlisting}

\begin{lstlisting}[style=json, caption={$G_4$: Dual-wavelength reflective design.}, label={lst:task-g4}]
{
  (*@\textcolor{jsonkey}{"query"}@*): "Design a reflective grating jointly for 0.632 um and 0.530 um. Use period 0.4777 um, high-index layer n=2.436, substrate n=1.363, and TE polarization. Optimize geometry and thickness to maximize the product of reflection at both wavelengths.",
  (*@\textcolor{jsonkey}{"gt\_eval"}@*): {
    (*@\textcolor{jsonkey}{"wavelength\_um"}@*): [0.53, 0.632],
    (*@\textcolor{jsonkey}{"criteria"}@*): [
      {(*@\textcolor{jsonkey}{"metric"}@*): "total_reflection", (*@\textcolor{jsonkey}{"params"}@*): {(*@\textcolor{jsonkey}{"wavelength\_index"}@*): 0},
       (*@\textcolor{jsonkey}{"operation"}@*): ">=", (*@\textcolor{jsonkey}{"target"}@*): 0.7},
      {(*@\textcolor{jsonkey}{"metric"}@*): "total_reflection", (*@\textcolor{jsonkey}{"params"}@*): {(*@\textcolor{jsonkey}{"wavelength\_index"}@*): 1},
       (*@\textcolor{jsonkey}{"operation"}@*): ">=", (*@\textcolor{jsonkey}{"target"}@*): 0.7}
    ]
  },
  (*@\textcolor{jsonkey}{"reference"}@*): "10.1038/s42005-021-00568-6"
}
\end{lstlisting}

\begin{lstlisting}[style=json, caption={$G_5$: Plasmonic reflective design.}, label={lst:task-g5}]
{
  (*@\textcolor{jsonkey}{"query"}@*): "Design a plasmonic reflective grating at 0.632 um. Use period 0.3779 um, metallic patterned layer with n=1.3523, k=7.9137, and substrate n=1.363. Use TE polarization at normal incidence and maximize total reflection efficiency.",
  (*@\textcolor{jsonkey}{"gt\_eval"}@*): {
    (*@\textcolor{jsonkey}{"wavelength\_um"}@*): [0.632],
    (*@\textcolor{jsonkey}{"criteria"}@*): [
      {(*@\textcolor{jsonkey}{"metric"}@*): "total_reflection", (*@\textcolor{jsonkey}{"params"}@*): {(*@\textcolor{jsonkey}{"wavelength\_index"}@*): 0},
       (*@\textcolor{jsonkey}{"operation"}@*): ">=", (*@\textcolor{jsonkey}{"target"}@*): 0.8}
    ]
  },
  (*@\textcolor{jsonkey}{"reference"}@*): "10.1038/s42005-021-00568-6"
}
\end{lstlisting}

\begin{lstlisting}[style=json, caption={$G_6$: Transmissive phase-target design (PbTe/CaF\textsubscript{2} rectangular pillar).}, label={lst:task-g6}]
{
  (*@\textcolor{jsonkey}{"query"}@*): "Design a rectangular pillar metasurface unit cell at
    5.200 um. The grating layer is 638 nm total thickness and modeled as two PbTe layers (top half: n=4.498, k=0.000; bottom half: n=5.937, k=0.010). The substrate is CaF2 with n=1.272, and the superstrate is air. Periodicity is 2.699 um in both x and y. Use TM polarization at normal incidence, transmitted from substrate to air. Optimize pillar length and width to exceed 0.56 total transmission and reach a transmitted TM phase of 347.8 degrees with phase error < 5.0 degrees.",
  (*@\textcolor{jsonkey}{"gt\_eval"}@*): {
    (*@\textcolor{jsonkey}{"wavelength\_um"}@*): [5.2],
    (*@\textcolor{jsonkey}{"criteria"}@*): [
      {(*@\textcolor{jsonkey}{"metric"}@*): "total_transmission", (*@\textcolor{jsonkey}{"params"}@*): {(*@\textcolor{jsonkey}{"wavelength\_index"}@*): 0},
       (*@\textcolor{jsonkey}{"operation"}@*): ">=", (*@\textcolor{jsonkey}{"target"}@*): 0.5566},
      {(*@\textcolor{jsonkey}{"metric"}@*): "zero_order_transmission_phase_deg", (*@\textcolor{jsonkey}{"params"}@*): {(*@\textcolor{jsonkey}{"component"}@*): "x", (*@\textcolor{jsonkey}{"wavelength\_index"}@*): 0},
       (*@\textcolor{jsonkey}{"operation"}@*): "close_to", (*@\textcolor{jsonkey}{"target"}@*): 347.7508, (*@\textcolor{jsonkey}{"tolerance"}@*): 5.0}
    ]
  },
  (*@\textcolor{jsonkey}{"reference"}@*): "10.1515/nanoph-2025-0507"
}
\end{lstlisting}

\begin{lstlisting}[style=json, caption={$G_{\mathrm{aux}}$: Transmissive phase-target design{,} TiO\textsubscript{2} on SiO\textsubscript{2} glass pillar variant.}, label={lst:task-gaux}]
{
  (*@\textcolor{jsonkey}{"query"}@*): "Design a rectangular TiO2 pillar metasurface unit cell at 0.480 um wavelength. The pillar is made of TiO2 on an SiO2 glass substrate with air superstrate. Periodicity is 0.3729 um in both x and y. The pillar height is 558 nm. Use TE polarization at normal incidence, transmitted from substrate to air. Optimize pillar length and width to exceed 0.75 total transmission and reach a transmitted TE phase of 359.7 degrees with phase error < 5.0 degrees.",
  (*@\textcolor{jsonkey}{"gt\_eval"}@*): {
    (*@\textcolor{jsonkey}{"wavelength\_um"}@*): [0.48],
    (*@\textcolor{jsonkey}{"criteria"}@*): [
      {(*@\textcolor{jsonkey}{"metric"}@*): "total_transmission", (*@\textcolor{jsonkey}{"params"}@*): {(*@\textcolor{jsonkey}{"wavelength\_index"}@*): 0},
       (*@\textcolor{jsonkey}{"operation"}@*): ">=", (*@\textcolor{jsonkey}{"target"}@*): 0.753},
      {(*@\textcolor{jsonkey}{"metric"}@*): "zero_order_transmission_phase_deg", (*@\textcolor{jsonkey}{"params"}@*): {(*@\textcolor{jsonkey}{"component"}@*): "y", (*@\textcolor{jsonkey}{"wavelength\_index"}@*): 0},
       (*@\textcolor{jsonkey}{"operation"}@*): "close_to", (*@\textcolor{jsonkey}{"target"}@*): 359.7388, (*@\textcolor{jsonkey}{"tolerance"}@*): 5.0}
    ]
  },
  (*@\textcolor{jsonkey}{"reference"}@*): "10.1364/OE.27.030308"
}
\end{lstlisting}

\section{Training Dynamics}
\label{app:agentic-training-dynamics}

\begin{table}[htbp]
    \centering
    \caption{Training and validation dynamics across skill-evolution iterations, for the three settings.}
    \label{tab:agentic-training-dynamics}
    \small
    \setlength{\tabcolsep}{4pt}
    \begin{tabular}{lcccccc}
    \toprule
     & \multicolumn{2}{c}{Same-type} & \multicolumn{2}{c}{New-type-A} & \multicolumn{2}{c}{New-type-B} \\
    \cmidrule(lr){2-3}\cmidrule(lr){4-5}\cmidrule(lr){6-7}
    Row & SG (\%) & CPF & SG (\%) & CPF & SG (\%) & CPF \\
    \midrule
    Starter-skill baseline (test) & 38 & 0.51 & 92 & 0.96 & 20 & 0.34 \\
    Iteration 1 (train / val) & 48 / 100 & 0.58 / 1.00 & 52 / 100 & 0.61 / 1.00 & 92 / 13 & 0.96 / 0.23 \\
    Iteration 2 (train / val) & 80 / 100 & 0.89 / 1.00 & 48 / 93 & 0.62 / 0.93 & 92 / 100 & 0.96 / 1.00 \\
    Iteration 3 (train / val) & 80 / 100 & 0.90 / 1.00 & 42 / 80 & 0.44 / 0.80 & 92 / 100 & 0.96 / 1.00 \\
    Iteration 4 (train / val) & 80 / 100 & 0.87 / 1.00 & 26 / 100 & 0.34 / 1.00 & 90 / 100 & 0.95 / 1.00 \\
    Post-train (test) & 74 & 0.87 & 90 & 0.95 & 90 & 0.93 \\
    \bottomrule
    \end{tabular}
\end{table}

Table~\ref{tab:agentic-training-dynamics} reports the training and validation metrics of the candidate skill files considered during skill evolution. The starter-skill baseline and post-train rows are test-set evaluations before and after evolution. The intermediate rows summarize the training and validation outcomes at each iteration. Validation metrics provide the selection signal, while training metrics reflect adaptation on the sampled batch of training tasks.

The table is not a strict adjacent-iteration lineage. As described in the main text, each iteration starts from the best previously selected skill file according to validation performance. The trajectory therefore shows successive iterations, each starting from the best skill file selected so far. The post-train row is evaluated on the fixed held-out test split rather than on the validation split used for selection, which makes it directly comparable to the starter-skill baseline row.

The table mirrors the main results. Under \SameType, validation SG reaches 100\% early and training CPF rises substantially, consistent with rapid improvement within seen template families. Under \NewTypeA, the starter is already near ceiling on the test families, so validation success peaks at iteration 1 and the selected skill leaves test performance essentially unchanged, with a net regression of one task. Under \NewTypeB, validation SG stays low at iteration~1 (13\%) and then reaches 100\% at iteration~2, which is the selected iteration; correspondingly, the post-train test SG rises to 90\% from the 20\% starter-skill baseline.

\section{Agent Tool-Use Behavior}
\label{app:agentic-tool-use}

\begin{figure}[htb!]
    \centering
    \includegraphics[width=\textwidth]{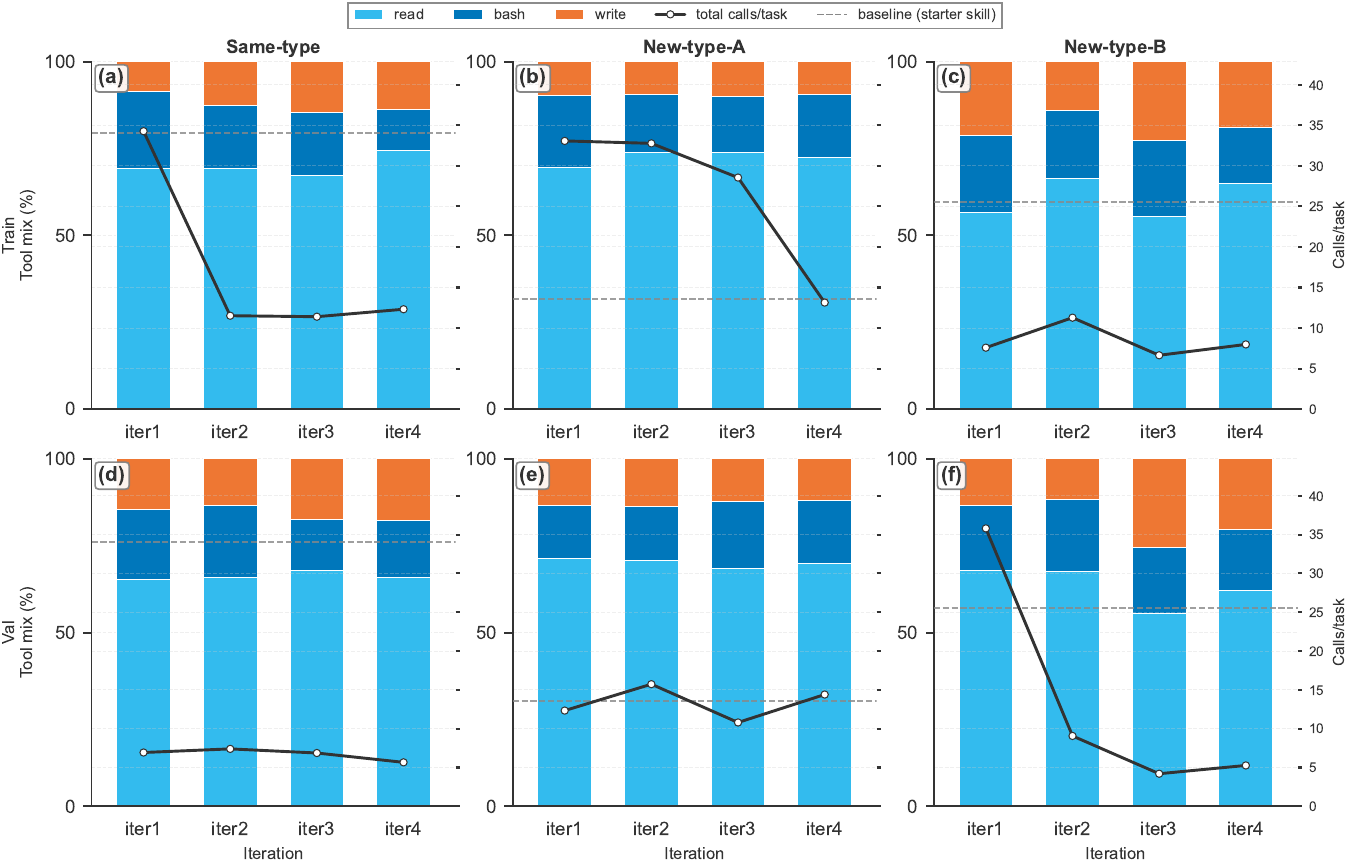}
    \caption{Tool-call composition during skill evolution, by setting (columns: \SameType, \NewTypeA, \NewTypeB) and split (top row: training; bottom row: validation). Stacked bars give the per-task share of read, bash, and write tool calls at each iteration; the black line is the total calls per task (right axis) and the dashed line is the fixed starter-skill baseline. Read and bash calls correspond to exploration and write calls to revision, so a shift toward shorter, write-led cycles indicates less effort spent on broad search as the skill matures.}
    \label{fig:si-tool-composition}
\end{figure}

Figure~\ref{fig:si-tool-composition} reports how the coding agent allocates its tool budget per task across skill-evolution iterations, with the training split in the top row and the validation split in the bottom row of each setting. Read and bash calls correspond to exploration and write calls to revision. On \SameType, later iterations move away from broad read and bash exploration toward shorter, write-led revision cycles, with the total calls per task falling below the starter-skill baseline. The new-type settings show smaller and less systematic shifts in the tool mix, consistent with held-out families still requiring task-by-task probing rather than a single reusable search pattern.

\section{Error Taxonomy}
\label{app:agentic-error-taxonomy-code}

\begin{table}[H]
    \centering
    \caption{Error categories used in the main error-analysis figure.}
    \label{tab:agentic-error-taxonomy}
    \begin{tabular}{lp{10.2cm}}
        \toprule
        Category & Definition \\
        \midrule
        Tensor Index OOB & Errors caused by invalid tensor indexing, shape mismatch, or out-of-bounds tensor access during candidate-program execution. \\
        API Misuse & Errors caused by incorrect use of solver or optimization APIs, including invalid argument conventions and unsupported call patterns. \\
        Gradient Error & Failures related to broken computational graphs, missing gradients, or invalid backward propagation through the optimization objective. \\
        No Code / No Solution & Cases in which the Coding Agent fails to produce a complete executable candidate, or produces a candidate that never reaches a valid solution. \\
        Excluded infrastructure failures & Connection errors, timeouts, and similar runtime-environment failures are excluded from Figure~\ref{fig:mechanism} because they are not directly attributable to the learned skill content. \\
        \bottomrule
    \end{tabular}
\end{table}

\section{Illustrative Evolution of the Skill File}
\label{app:agentic-skill-case-study}

\newcounter{skilllisting}[section]
\renewcommand{\theskilllisting}{\thesection.\arabic{skilllisting}}

This section provides a qualitative case study of the learned skill file used in the main text. Rather than presenting the full evolution trajectory across all iterations, we compare two representative snapshots from the same-type run: a starter skill and the selected evolved skill at iteration~1. The purpose of this appendix is to make the learned skill file directly inspectable and to illustrate how the evolved skill moves beyond a minimal {\solver} setup guide toward a more structured set of rules for task classification, failure avoidance, and reusable strategy selection.

Table~\ref{tab:agentic-skill-case-study-summary} summarizes the two snapshots included below. The starter skill is a lightweight file that primarily encodes solver setup and minimal execution patterns. The evolved skill corresponds to the archived iteration-1 snapshot selected during the main training trajectory and therefore provides a representative example of the qualitative transition induced by skill evolution.

\begin{table}[htbp]
    \centering
    \caption{Representative skill snapshots used in the qualitative case study.}
    \label{tab:agentic-skill-case-study-summary}
    \small
    \setlength{\tabcolsep}{4pt}
    \begin{tabular}{p{2.3cm}p{4.1cm}p{1.7cm}p{5.2cm}}
        \toprule
        Snapshot & Source & Length & Primary role \\
        \midrule
        starter skill & same-type run, starter-skill snapshot (iter0) & 92 lines & Minimal TorchRDIT setup and execution pattern \\
        Evolved skill & same-type run, archived selected snapshot (iter1) & 218 lines & Structured task classification, failure-aware guidance, and reusable strategy selection \\
        \bottomrule
    \end{tabular}
\end{table}

\refstepcounter{skilllisting}\label{lst:agentic-skill-iter0}
\begin{MarkdownBlock}{Listing~\theskilllisting\ \textbar\ starter skill file (iter0)}
---
name: learning-context
description: Minimal TorchRDIT bootstrap plus navigation for inverse design.
---

# TorchRDIT Inverse Design Bootstrap

## Skill Overview

Start from the minimal TorchRDIT contract in this file. Keep imports canonical, build a solver with `get_solver_builder()`, create the source with `solver.add_source(...)`, and return the direct `solver.solve(...)` result. Read `reference/*.md` only when the query or a runtime error requires deeper context.

## Bootstrap Rules

- Implement `def solve_inverse_design(*, device: str = "cpu") -> SolverResults`.
- Import TorchRDIT symbols from the documented paths in this file; do not invent modules.
- Register materials before they are referenced by name in layers or boundary media.
- Add finite layers once during setup, then modify them in place inside optimization loops.
- Always call `.with_device(device)` early in the builder chain to ensure all solver tensors are placed on the correct compute device.
- Build the source with `solver.add_source(...)` and return the direct result of `solver.solve(source)`.


## Required Imports

```python
import numpy as np
import torch

from torchrdit.constants import Algorithm, Precision
from torchrdit.results import SolverResults
from torchrdit.solver import get_solver_builder
from torchrdit.utils import create_material
```

## Builder Setup

```python
builder = get_solver_builder()
builder.with_device(device)
builder.with_algorithm(Algorithm.RCWA)
builder.with_precision(Precision.DOUBLE)
builder.with_wavelengths(np.array([1.55]))
builder.with_length_unit("um")
builder.with_real_dimensions([512, 512])
builder.with_k_dimensions([9, 9])
solver = builder.build()
```

## Materials And Layers

```python
air = create_material(name="air", permittivity=1.0)
film = create_material(name="film", permittivity=11.7)
solver.add_materials(material_list=[air, film])
solver.update_ref_material("air")
solver.add_layer(
     material_name="film",
    thickness=torch.tensor(0.22, dtype=torch.float64, device=device),
    is_homogeneous=True,
)
solver.update_trn_material("air")
```

## Source And Solve Pattern

```python
degrees = np.pi / 180
source = solver.add_source(theta=0 * degrees, phi=0 * degrees, pte=1.0, ptm=0.0)
result = solver.solve(source)
transmission_total = result.transmission
reflection_total = result.reflection
tx, ty, tz = result.get_zero_order_transmission()
```

## Reference Map

- `reference/setup.md` - imports, units, and builder configuration.
- `reference/materials_and_layers.md` - materials, boundary media, and stack ordering.
- `reference/patterning.md` - mask construction and patterned layers.
- `reference/sources_and_solving.md` - source setup, batched solves, and execution.
- `reference/phase_and_orders.md` - phase extraction and diffraction-order reads.
- `reference/amplitude_and_efficiency.md` - amplitude and efficiency metrics.
- `reference/optimization.md` - deterministic optimization recipes.
- `reference/pitfalls.md` - high-frequency failure modes and fixes.
- `reference/context_usage.md` - how to use copied context artifacts.

## Strategy Portfolio

Pick one bounded global search family before coding, then refine locally only after a promising candidate appears. Good default families are: multistart/global screening, small population-style candidate search, or coarse-to-fine parameter sweeps. Keep the first line of `solution.py` as `# Strategy: <family> - <why>` so retries can preserve or change the family deliberately instead of thrashing.

## Optional Context

If a `context/` directory exists, treat it as optional learned guidance. Read `context/strategy_summary.md` first when present. It is the cross-iteration memory mirrored from `meta_agent/strategy_summary.md`, so use it to anchor durable strategy choices before consulting batch-local context. Then read only the most relevant remaining context files after `SKILL.md` and the needed `reference/*.md` files.

\end{MarkdownBlock}

\refstepcounter{skilllisting}\label{lst:agentic-skill-iter1}
\begin{MarkdownBlock}{Listing~\theskilllisting\ \textbar\ Evolved skill file (iter1)}
---
name: learning-context-iter1
description: "Learning-Context Skill: Inverse Design Code Generation"
---

# Learning-Context Skill: Inverse Design Code Generation

## Skill Overview

This skill guides the base agent (context engineer) to learn from training data and produce high-quality `solve_inverse_design()` Python functions. The agent reads evaluation results, extracts patterns from successes and failures, maintains structured context, and incrementally improves solution quality across sub-iterations.

The goal is to satisfy authoritative dataset-owned criteria constraints (transmission efficiency, phase targets, diffraction orders) by writing correct, robust TorchRDIT-based inverse-design code.

---

## Methodology

### Step 1: Load Prior Context

Read all existing files in `context/` before doing anything else:

```
context/rules.md           # Accumulated solution rules and patterns
context/examples.md        # Working solution templates with annotations
context/failure_modes.md   # Known failure patterns to avoid
context/strategy_summary.md  # Durable strategic lessons (read-only mirror)
```

If a file doesn't exist yet, create it in Step 4 after analysis.

### Step 2: Load and Parse Training Results

Load `data/train.json`. For each sample, extract:
- `sample_id`, `query` (the task description)
- `criteria` list (what must be satisfied)
- `success` / `criteria_pass_fraction` / `criteria_violation_norm`
- `best_margin` (minimum normalized criterion margin for the task)
- `error_reason` if execution failed
- The generated `solution.py` content (if available in per-task traces)

Group samples into:
- **Full successes**: `success == true` (all criteria pass)
- **Partial successes**: `criteria_pass_fraction > 0` but `success == false`
- **Full failures**: `criteria_pass_fraction == 0` or execution error

### Step 3: Analyze Patterns

**From successes**: Extract what structural choices (layer count, material, thickness range, optimizer type/steps, convergence) led to all criteria passing. Look for recurring patterns across multiple successful samples.

**From failures**: Identify the top failure signatures:
- Execution errors (import failure, shape mismatch, device errors)
- Criteria-specific misses (which criterion fails most often — transmission vs. phase vs. diffraction order)
- Optimizer stagnation: did the loss plateau without satisfying constraints?

**Key diagnostic questions**:
1. Which criteria types fail most often? (e.g., phase accuracy vs. transmission magnitude)
2. Are failures execution errors or optimization shortfalls?
3. Do successes share a common structural pattern (number of layers, optimizer, step count)?
4. Is there a near-miss pattern (close but not satisfying tolerance)?

### Step 4: Update Context Files

Update `context/` files based on evidence. Keep changes targeted and evidence-based:

**`context/rules.md`**: Actionable rules derived from failure analysis:
```markdown
## Rules (Evidence-Based)
- Always call `.with_device(device)` early in the builder chain.
- Transfer tensors to CPU before NumPy export: `.detach().cpu().numpy()`
- For phase criteria: use sufficient optimization steps (≥ 100) with fine-grained lr.
- For transmission criteria: ensure the solver wavelength list matches `gt_eval` wavelength_um.
- For multi-criteria tasks: use a composite loss that weights all criteria terms.
- After optimization: verify result passes `SolverResults` interface before returning.
```

**`context/examples.md`**: Annotated working solution templates. Store the simplest full-success example for each criteria type observed in training data.

**`context/failure_modes.md`**: Failure fingerprints to avoid:
```markdown
## Failure Modes
- Forgetting `.detach()` before `.numpy()` → RuntimeError
- Using wrong wavelength index in criteria → always match wavelength_um list order
- Flat loss (optimizer stagnated) → add random restarts or larger initial perturbation
- Shape mismatch in permittivity tensor → check layer/grid dimension conventions
```

### Step 5: Write the Solution for Each Training Sample

For each sample in the current batch (task-environment will call you once per sample), write `solution.py` that:

1. **Reads the query** to understand what the task requires (geometry type, material, wavelength, target criteria).
2. **Applies rules from `context/rules.md`** to avoid known failure modes.
3. **Uses templates from `context/examples.md`** as starting points.
4. **Implements `solve_inverse_design(*, device: str = "cpu") -> SolverResults`** with:
   - Correct builder chain (device placement early)
   - Appropriate optimization loop (enough steps, sensible lr schedule)
   - Composite loss covering ALL criteria in the task
   - CPU transfer of all tensors before returning

5. **Adds a `# Strategy:` comment near the top** summarizing which patterns from context you applied and why, for meta-agent feedback.

### Step 6: Reflect and Finalize Context

After writing solutions for all samples in the batch:
- Note which rules were most frequently applied.
- Flag any new failure pattern not yet in `context/failure_modes.md`.
- If a new success pattern emerges, add a concise example to `context/examples.md`.
- Commit updated context files so they carry forward to the next sub-iteration.

---

## TorchRDIT API Reference (Compact)

Below are the key patterns for calling TorchRDIT's builder/solver APIs. Adapt to the specific task.

### Minimal Builder Pattern

```python
from torchrdit.solver import create_solver
from torchrdit.builder import SolverBuilder

def solve_inverse_design(*, device: str = "cpu") -> "SolverResults":
    import torch

    builder = (
        SolverBuilder()
        .with_device(device)                     # always first
        .with_wavelengths([1.55])                # match gt_eval wavelength_um
        .with_materials({"Si": 3.5, "SiO2": 1.5})
        .with_layer_structure([...])             # task-specific
        .with_source(...)                        # task-specific
    )
    solver = builder.build()

    # Optimization loop
    params = torch.nn.Parameter(torch.rand(..., device=device))
    optimizer = torch.optim.Adam([params], lr=1e-2)

    for step in range(200):
        optimizer.zero_grad()
        results = solver.solve(params)
        loss = compute_loss(results)             # covers all criteria
        loss.backward()
        optimizer.step()

    # Final solve with detached params
    with torch.no_grad():
        final_results = solver.solve(params)

    # Transfer to CPU
    return final_results.to_cpu()               # or manual .detach().cpu()
```

### Composite Loss for Multiple Criteria

```python
def compute_loss(results):
    loss = torch.tensor(0.0, device=results.device)

    # Transmission criterion: target >= 0.8
    T = results.transmission[0]  # wavelength index 0
    loss += torch.relu(0.8 - T)

    # Phase criterion: target 170° ± 8.5° (use circular distance)
    phase = results.phase_deg[0]  # component x, wavelength 0
    diff = ((phase - 170.0 + 180) 
    loss += torch.relu(torch.abs(diff) - 8.5)

    return loss
```

### Device and Tensor Safety

```python
# Always end with CPU transfer
result = result.detach().cpu()  # for individual tensors
# OR use built-in method if available
return solver_results  # ensure internal tensors are .detach().cpu() before NumPy
```

---

## Optimization Strategy Guidance

These are starting heuristics; adjust based on evidence from training data:

| Scenario | Recommended Strategy |
|---|---|
| Single transmission criterion | Adam, lr=1e-2, 150 steps |
| Single phase criterion | Adam, lr=5e-3, 300 steps (phase is sensitive) |
| Multi-criteria (T + phase) | Adam, composite loss, 300–500 steps; lr warmup |
| Optimizer stagnation | Random restart: reinitialize params, run 100 more steps |
| Near-miss (close to tolerance) | Fine-tune: lr=1e-4, 100 additional steps |

**Global exploration before local refinement**: If training shows high `optimizer_limited_rate`, add a coarse random search (5–10 random initializations, pick best, then refine) before running the main optimization.

---

## Context Engineering Principles

1. **Build incrementally**: Always start from existing context; never overwrite blindly.
2. **Evidence-first**: Only add rules/examples supported by actual training data results.
3. **Specificity beats generality**: Concise, specific rules outperform vague guidelines.
4. **Failure fingerprints are gold**: Document exact error messages and their fixes.
5. **Criteria authority**: Do not alter target values — satisfy them as given.
6. **Strategy transparency**: The `# Strategy:` comment in each `solution.py` enables meta-agent learning.

---

## Output Checklist

Before ending each sub-iteration:
- [ ] `context/rules.md` updated with new evidence-based rules
- [ ] `context/examples.md` has at least one annotated success template
- [ ] `context/failure_modes.md` lists top observed failure patterns
- [ ] Each `solution.py` has a `# Strategy:` comment
- [ ] All solutions use `.with_device(device)` and `.detach().cpu()` patterns
- [ ] Loss function covers ALL criteria in the task query

\end{MarkdownBlock}

Comparing Listings~\ref{lst:agentic-skill-iter0} and~\ref{lst:agentic-skill-iter1} shows three concrete changes. First, the evolved skill file is much more operational: it moves beyond a starter API guide and spells out how to read prior context, parse training results, group task types, and update reusable guidance. Second, it turns repeated failure patterns into explicit procedural rules. Third, it becomes more task-aware, moving from generic setup instructions toward a more specialized, evidence-conditioned protocol. These qualitative changes complement the quantitative results in the main text by showing that the learned object remains explicit and inspectable rather than hidden in model parameters.

\end{appendices}

\end{document}